\documentclass[runningheads]{llncs}

\usepackage{eccv}

\usepackage{eccvabbrv}

\usepackage{graphicx}
\usepackage{booktabs}

\usepackage[accsupp]{axessibility}  

\usepackage{hyperref}

\usepackage{orcidlink}

\usepackage{multirow}
\usepackage{booktabs}
\usepackage{xcolor}
\usepackage{float}
\usepackage{makecell}

\begin{document}

\title{Aligning Object Detector Bounding Boxes\\with Human Preference} 

\titlerunning{Aligning Object Detector Bounding Boxes with Human Preference}

\author{Ombretta Strafforello\inst{1,2}\orcidlink{0000-0002-5258-8534} 
\and
Osman S. Kayhan\inst{1,3}\orcidlink{0000-0002-0328-7647} 
\and
Oana Inel\inst{4}\orcidlink{0000-0003-4691-6586}
\and
Klamer Schutte\inst{2}\orcidlink{0000-0002-9954-0685} 
\and
Jan van Gemert\inst{1}\orcidlink{0000-0002-3913-2786}}

\authorrunning{O.~Strafforello et al.}

\institute{Delft University of Technology, Delft, the Netherlands \\ 
\and
TNO, The Hague, the Netherlands \\ 
\and
Haga Teaching Hospital, The Hague, the Netherlands \\ 
\and
University of Zurich, Zurich, Switzerland \\
\email{\{o.strafforello, o.s.Kayhan, j.c.vangemert\}@tudelft.nl}, \\
\email{klamer.schutte@tno.nl}, 
\email{inel@ifi.uzh.ch}}

\maketitle

\begin{abstract}

Previous work shows that humans tend to prefer large bounding boxes over small bounding boxes with the same IoU. However, we show here that commonly used object detectors predict large and small boxes equally often. In this work, we investigate how to align automatically detected object boxes with human preference and study whether this improves human quality perception. We evaluate the performance of three commonly used object detectors through a user study (N = 123). We find that humans prefer object detections that are upscaled with factors of 1.5 or 2, even if the corresponding AP is close to 0.
Motivated by this result, we propose an asymmetric bounding box regression loss that encourages large over small predicted bounding boxes. 
Our evaluation study shows that object detectors fine-tuned with the asymmetric loss are better aligned with human preference and are preferred over fixed scaling factors. A qualitative evaluation shows that human preference might be influenced by some object characteristics, like object shape. 

\end{abstract}

\section{Introduction}
\label{sec:introduction}

Object detectors identify and localize objects in an image. We focus on the common setting where detections are presented to a human by drawing a bounding box around the objects. In this paper, we evaluate how to best present object detections to humans, which is paramount for all applications that rely on showing detection to humans, such as visual inspection~\cite{kayhan2021hallucination, li2013rail, mery2017logarithmic}, anomaly detection~\cite{basharat2008learning, doshi2020fast, li2018object}, or medical imaging~\cite{li2019clu, litjens2017survey}. 
%
Previous work showed that humans prefer larger over smaller boxes with the same localization error~\cite{strafforello2022humans}. This was concluded in an online study with a fully controlled setup, where ground truth bounding boxes are precisely matched to the localization error. 
However, it is not directly clear if this controlled setting translates to the real world, where object detector outputs are imperfect.
In this work, we extend~\cite{strafforello2022humans} to real-world settings and real object detectors, which is important for reproducibility, and realistic, practical applications of scientific results.

\begin{figure}[!ht]
\centering
\includegraphics[width=0.9\columnwidth]{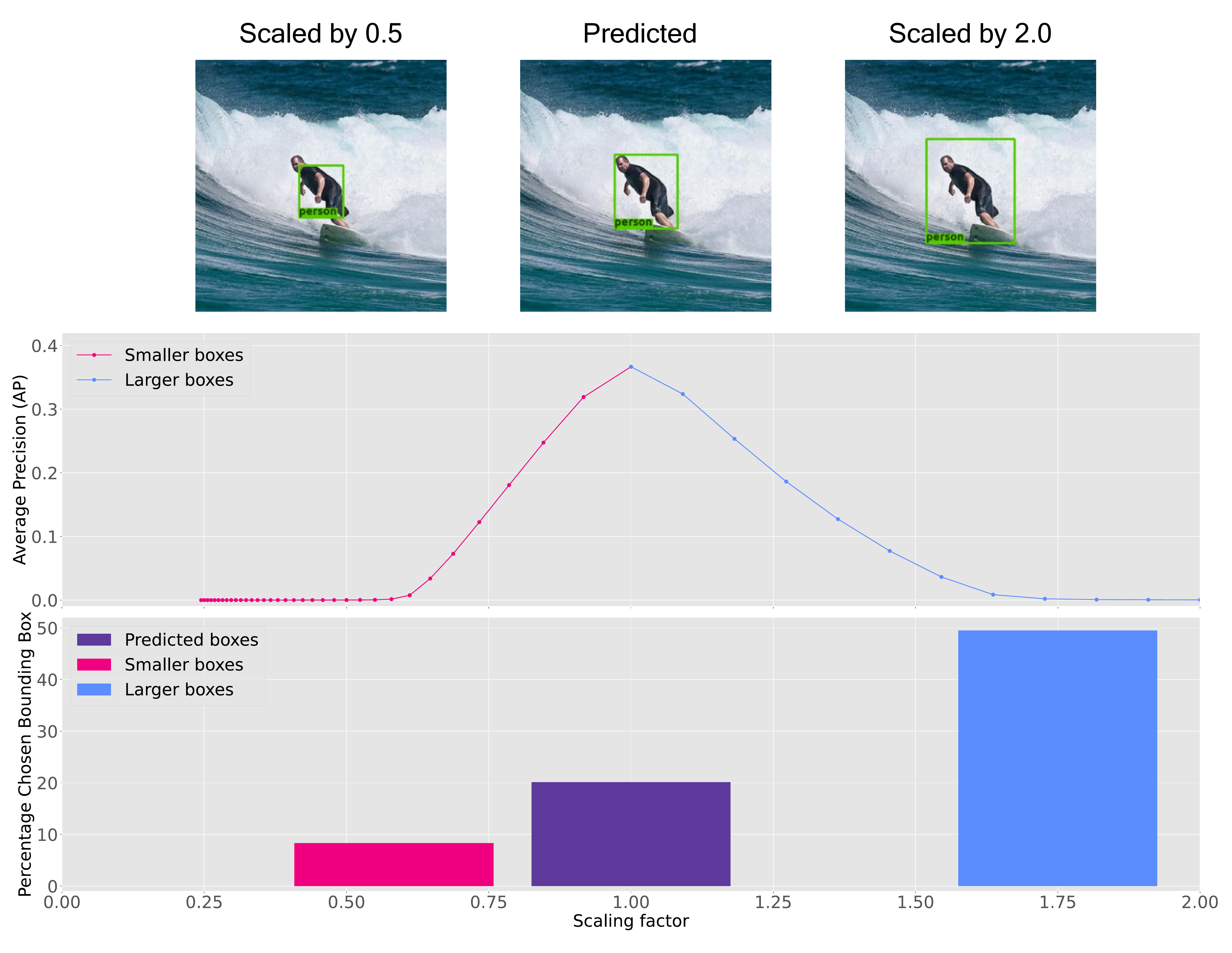}
\caption{
Scaling the predicted bounding box of Faster R-CNN~\cite{ren2015faster} on the COCO~\cite{lin2014microsoft} validation set. Average Precision (AP) (\textit{top}) versus human preference (\textit{bottom}). A scaling factor of $1.0$ corresponds to the original bounding box size. Upscaling and downscaling the size of the bounding boxes severely deteriorates AP. However, our study shows that humans prefer larger bounding boxes, even if they give nearly 0 AP.
}
\label{fig:AP_vs_humans}
\end{figure}

Object detectors, such as two-stage~\cite{cai2018cascade, ren2015faster}, single stage~\cite{lin2017focal,ssd, yolo}, 
anchorless~\cite{duan2019centernet, law2018cornernet, zhou2019bottom}, and transformers-based detectors \cite{beal2020transformerbased, Carion_2020, Dai_2021_CVPR, zhu2020deformable} 
minimize a classification loss and a localization loss for bounding box fitting. 
The localization loss is symmetric for errors in bounding box size: a predicted box that is 10\% too large will give the same loss as a box that is 10\% too small. Here, we investigate how this symmetry affects human perception of object detections.

%
Object detectors are typically evaluated using average precision (AP)~\cite{everingham2010pascal, hoiem2012diagnosing, OpenImages}, which depends on the accuracy of the object classification and of the bounding box localization, as measured by the Intersection over Union (IoU) with the ground truth box.
%
We are not the first to reconsider object detection evaluation~\cite{chavali2016object, feng2021labels, hosang2015makes},
yet, those works all assume that a perfect-fitting bounding box is best. 
%
In contrast, here we investigate if \emph{a perfect-fitting bounding box may not be the best box} for presenting detections to humans.

We make the following contributions: \emph{(1)} We analyze three popular object detectors and find that they predict small and large bounding boxes equally often.
\emph{(2)} We analyze how humans perceive the predictions of the object detectors focusing on the bounding box size. As shown in Figure~\ref{fig:AP_vs_humans}, we find that humans prefer upscaled object detections, even with corresponding AP close to 0. \emph{(3)} We propose an asymmetric loss function that favors the prediction of large over small boxes. 
Our evaluation shows that fine-tuning with the asymmetric loss better aligns object detections with human preference. 
All our collected data, analyses, and code are available on GitHub\footnote{\url{https://github.com/ombretta/humans-vs-detectors}}.

\section{Related Work}
\label{sec:related_work}

\subsection{Presenting object detections to humans}

We take a nuanced view on evaluating object detection by identifying two distinct use-cases. Case~1: A bounding box is used as pre-processing for a follow-up algorithm such as instance segmentation~\cite{bolya2019yolact, he2017maskRCNN, Shen_2021_ICCV}, video object detection~\cite{Chen_2020_CVPR, han2021context, jiao2021new}, human pose estimation~\cite{cao2019openpose,Sun_2019_CVPR,xiao2018simple}, action recognition~\cite{ballan2021long}, etc. Case~2: A bounding box is drawn on the image, and the full image is presented directly to a human, with relevant use-cases such as visual inspection~\cite{kayhan2021hallucination, li2013rail, mery2017logarithmic}, anomaly detection~\cite{basharat2008learning, doshi2020fast, li2018object},  medical imaging~\cite{li2019clu, litjens2017survey}, etc. We argue that these two use-cases deserve different treatment. 
For case~1, where the bounding box is a pre-processing step, it is difficult to consider all possible follow-up algorithms, and a tightly fitting box around the object, as evaluated using IoU, seems reasonable. For case~2, however, the bounding box is the final end result and is shown to a human being. Case~2 allows directly evaluating the end result in user studies, to understand what humans actually prefer in their object detection. This is the focus of our paper.

% ...

% ...

\subsection{Evaluating object detectors}
All object detectors such as two-stage models \cite{cai2018cascade, girshick2015fast, girshick2015region, ren2015faster}, single stage approaches \cite{lin2017focal,ssd, yolo,redmon2018yolov3}, pointwise/anchorless methods \cite{duan2019centernet, law2018cornernet, zhou2019bottom}, and transformers-based detectors \cite{beal2020transformerbased, Carion_2020, Dai_2021_CVPR, zhu2020deformable} are commonly evaluated~\cite{pascal-voc-2012,hoiem2012diagnosing,OpenImages,Salton86IAP} with mean average precision: 
the mean of the per-class average precision scores. Average precision (AP) is the area under the precision-recall curve, created by ranking all detections by confidence, and then checking if a detection is correct according to the ground truth. The correctness of a detection depends on the classification: if the assigned class label is wrong, the detection is wrong. A second criterion for correctness is that the location and size of the detection have sufficient overlap with the ground truth box. 
For determining the overlap, the Intersection over Union (IoU) score $\frac{B_p \cap B_{gt}}{B_p \cup B_{gt}}$ is used, where $B_p$ is the predicted bounding box, and $B_{gt}$ is the ground truth bounding box. The location of a detection is correct if the IoU score is higher than a certain threshold, typically 0.5 or higher~\cite{pascal-voc-2012, lin2014microsoft}.
%
Usually, the reported AP corresponds to a specific IoU threshold, such as 0.50 (AP50), or the average across several IoU thresholds, such as AP@$[0.5:0.95]$. 

We are not the first to consider object detection evaluation~\cite{chavali2016object, feng2021labels, hosang2015makes, padilla2020survey, prasad2019object, sobti2021vmap}, yet, those works all assume that a predicted bounding box perfectly overlapping with the ground truth bounding box is best. In contrast, we here challenge the view that a best fitting bounding box is always best for presenting detections to humans. 
We base our challenge on the work of Strafforello~\etal~\cite{strafforello2022humans} who show in precisely controlled experiments on ground truth boxes that humans prefer larger boxes over smaller boxes.  
In this paper, we investigate the practical ramifications of  Strafforello~\etal~\cite{strafforello2022humans} by aligning real-world object detectors with human preference.  

\subsection{Optimizing object detectors}
Object detectors are typically optimized using an object classification loss and a bounding box regression loss for accurate localization, by aligning the IoU of the predicted box with the ground truth box. The regression loss, usually an L2 \cite{redmon2017yolo9000} or smoothed L1 \cite{liu2016ssd, ren2015faster} function, forces the box coordinates to be as close a possible to the ground truth, where the IoU is often optimized as an additional loss term~\cite{ren2015faster}.
Previous work proposed novel object detector losses to improve the accuracy, measured in AP. 
Examples include using the Absolute size IoU (AIoU) \cite{TIAN20221029} and the SCALoss \cite{zheng2022scaloss}. Other work designed a new loss term to achieve computational efficiency \cite{aswal2021designing}.
In our paper, we propose a simple asymmetric regression loss function that enhances the performance of object detectors with respect to human quality judgments. Previous work used asymmetric loss in Bayesian estimation \cite{basu1991bayesian} and for classification \cite{ridnik2021asymmetric}. To the best of our knowledge, we are the first to use an asymmetric loss for bounding box regression.

\subsection{Human annotations for object detection}
\label{sec:rw_human_annotations}

The adoption of crowdsourcing platforms such as Amazon Mechanical Turk \cite{mturk} or Prolific \cite{prolific} facilitated the collection of large training and testing datasets for computer vision tasks \cite{caba2015activitynet, di2013crowdsourcing,  krishna2017visual,lintott2008galaxy, russell2008labelme, yuen2009labelme}, in contrast to using in-house annotators~\cite{everingham2010pascal,xiao2010sun}. 
For object detection, crowdsourcing studies are extensively used to draw bounding boxes around objects that appear in images \cite{song2015robot, zhu2012we} and videos \cite{vondrick2013efficiently} and to draw the precise shape of the object \cite{russell2008labelme, yuen2009labelme}. 
To eliminate the need for clustering or averaging several bounding boxes for the same object, in~\cite{russakovsky2015imagenet, su2012crowdsourcing}, the authors proposed a three-step workflow, where one annotator performs one step: \emph{(1)} draws a bounding box around an object; \emph{(2)} validates the drawn bounding box and \emph{(3)} decides whether there are still objects that need to be annotated in the image. These steps are repeated until all objects in an image are annotated with bounding boxes. 
Experiments in which the crowd validates object detections showed that annotators tend to be lenient when validating bounding boxes, \emph{i.e.}, bounding boxes with IoU $<$ 0.5 are still accepted \cite{papadopoulos2016we}. 
Furthermore, analyses performed in \cite{russakovsky2015best} suggest that to efficiently and accurately localize all objects in an image, several crowdsourcing tasks are needed, such as verifying box correctness, verifying object presence, or naming the object. 


\begin{figure*}[t]
\centering
\begin{tabular}{c@{}c@{}c}
\includegraphics[width=0.33\textwidth]
{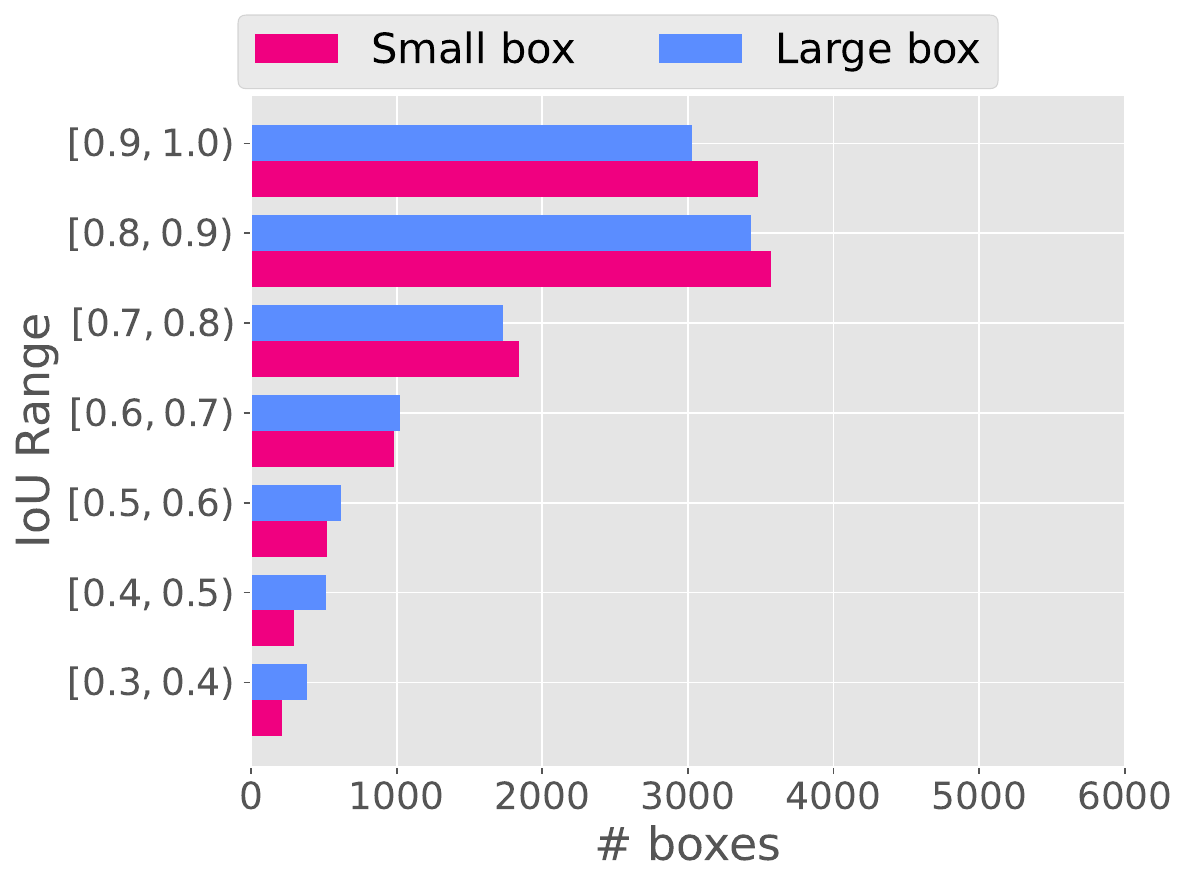} & 
\includegraphics[width=0.33\textwidth]{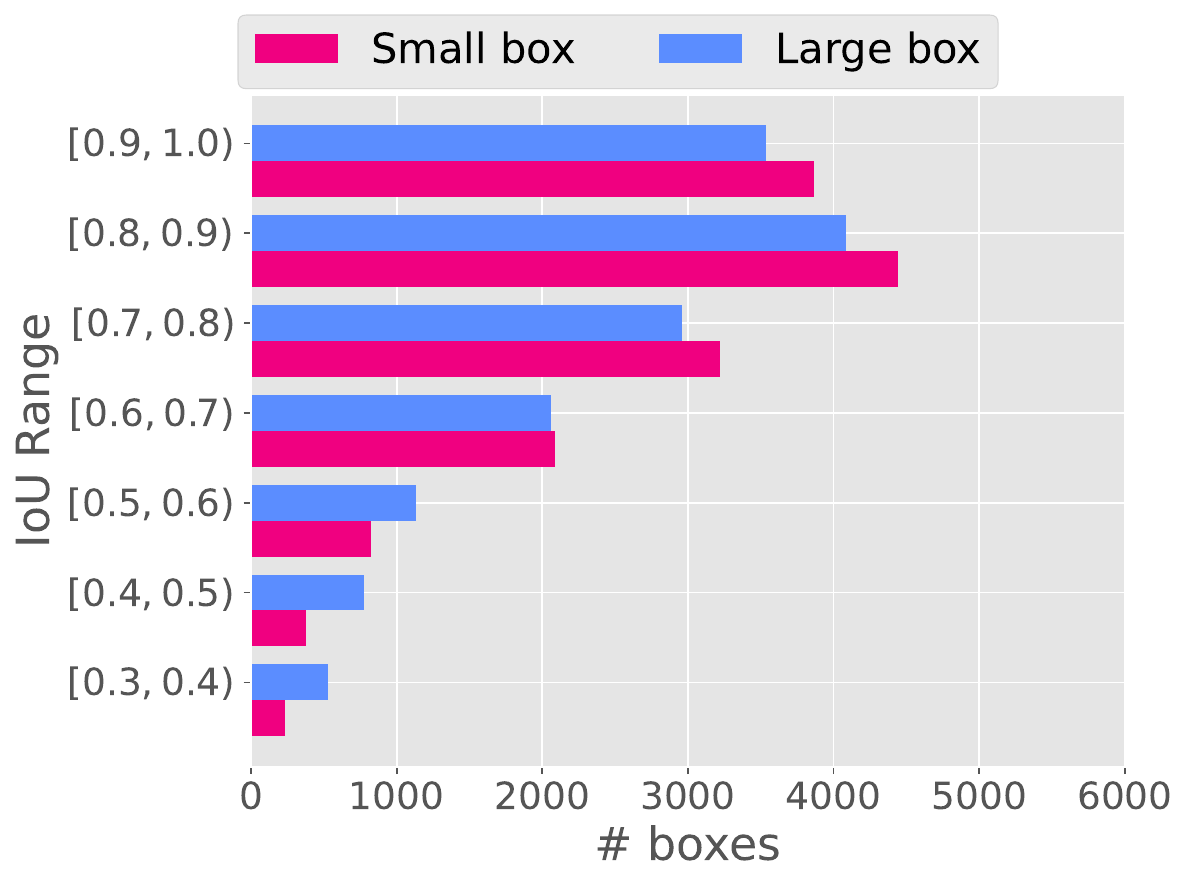} & 
\includegraphics[width=0.33\textwidth]{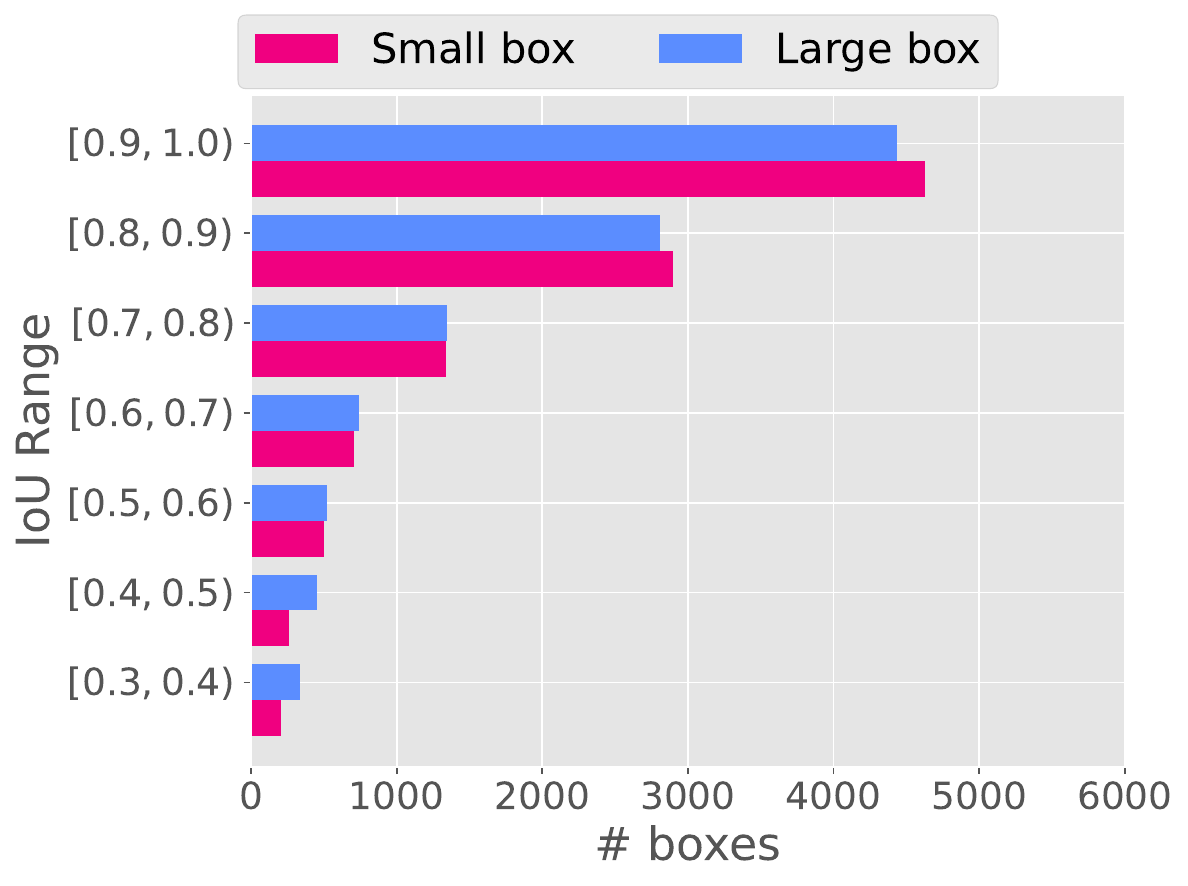}\\
(a) Faster R-CNN & (b) RetinaNet & (c) Cascade Mask R-CNN 
\end{tabular}
\caption{Amount of large and small bounding boxes are predicted by three object detectors on the MS COCO dataset, for seven IoU intervals, ranging from 0.3 to 1.0. 
For all three detectors, with higher IoU thresholds more small than large boxes are detected.
}
\label{fig:bboxes_sizes}
\end{figure*}


\section{Do humans prefer larger detections? }
\label{sec:experiment1}
Previous work shows that, for equal IoU, humans prefer too large boxes over too small boxes \cite{strafforello2022humans}. 
Here, we evaluate if this has practical consequences for real object detectors. We use three popular object detectors pretrained on MS COCO: Faster R-CNN \cite{ren2015faster}, RetinaNet \cite{lin2017focal}, and Cascade Mask R-CNN with ResNet-50 \cite{he2016deep} + Feature Pyramid Network \cite{lin2017feature} backbone \cite{cai2018cascade, DBLP:journals/corr/HeGDG17} all implemented in the Detectron2 library \cite{wu2019detectron2}.

\subsection{Do real detectors predict too large or too small boxes?}

If real object detectors tend to predict too large bounding boxes, then they are already well aligned with human preference. Thus, we investigate the relative size of the predicted bounding boxes with respect to the ground truth bounding box: A \textit{small box} has a smaller predicted area, and a \textit{large box} has a larger predicted area. We analyze predictions on the MS COCO validation set and count the occurrences of small and large boxes.

An overview of the distribution of the predicted bounding boxes over various IoU intervals is shown in Figure \ref{fig:bboxes_sizes}. For all object detectors that we examined, there is no statistically significant difference in the number of occurrences of large and small bounding boxes. This holds for small, medium, and large objects. However, for low IoU ranges, i.e., IoU $\in$ [0.3, 0.6) 
for Faster R-CNN and RetinaNet and IoU $\in$ [0.3, 0.5)  
for Cascade Mask R-CNN, large bounding boxes are more frequent than small ones. This is due to random 
large bounding boxes being more likely to partially overlap with the ground truth, compared to random 
small boxes. 
Considering intermediate IoU ranges, like IoU $\in$ [0.6, 0.7], 
the number of occurrences of small and large boxes is not in line with the human preference found in Strafforello \textit{et al.} \cite{strafforello2022humans}. That is, where humans would choose a large box over a small box with approximately 70\% chance, an object detector would predict small or large with nearby equal probability. 


We conclude that real object detectors generally do not predict too large boxes more often than too small boxes, and thus seem not well-aligned with human preference. In the following, we will investigate what this means for human quality judgments of real object detectors.  



\subsection{For real object detectors, do humans prefer too large boxes or too small boxes?}

\begin{figure*}[t]
\centering
\begin{tabular}{c@{}c@{}c@{}c@{}c}
\includegraphics[width=0.2\textwidth]{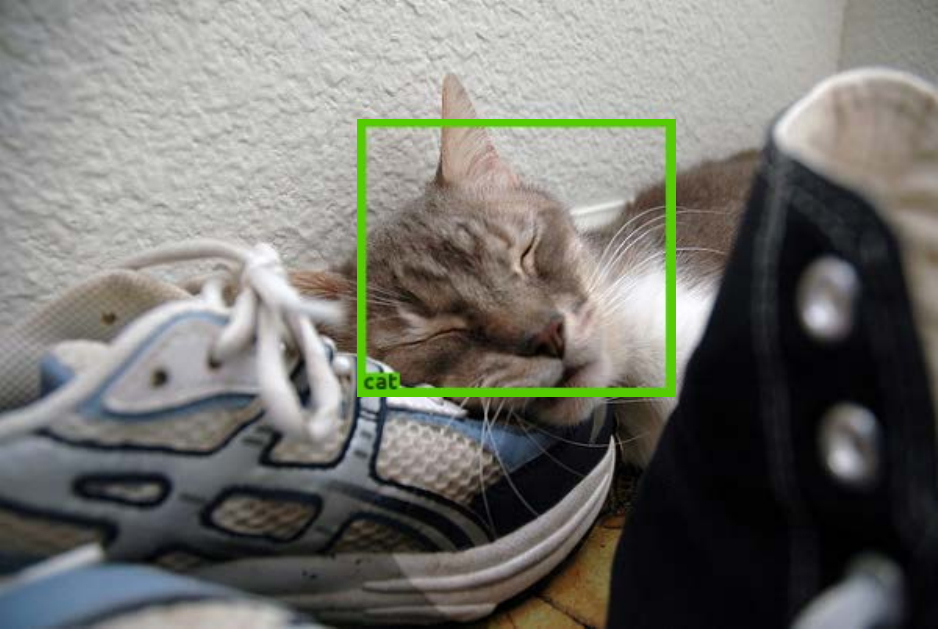} & 
\includegraphics[width=0.2\textwidth]{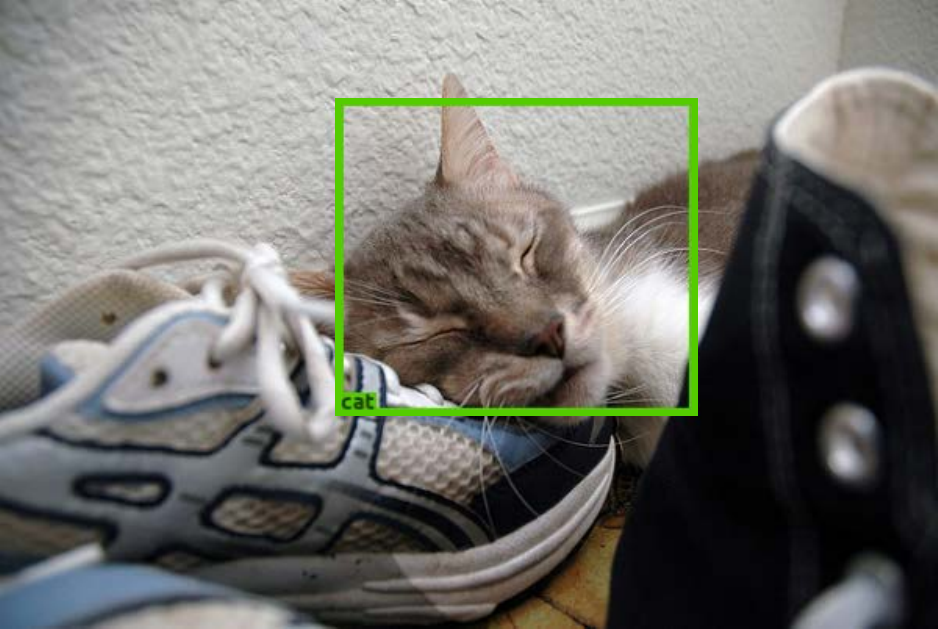} &
\includegraphics[width=0.2\textwidth]
{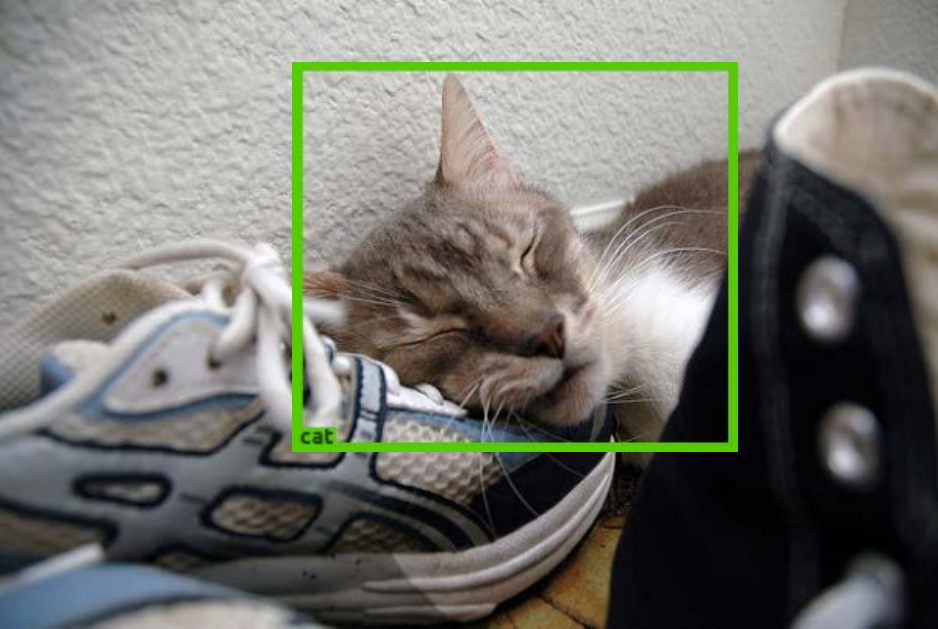} 
&
\includegraphics[width=0.2\textwidth]{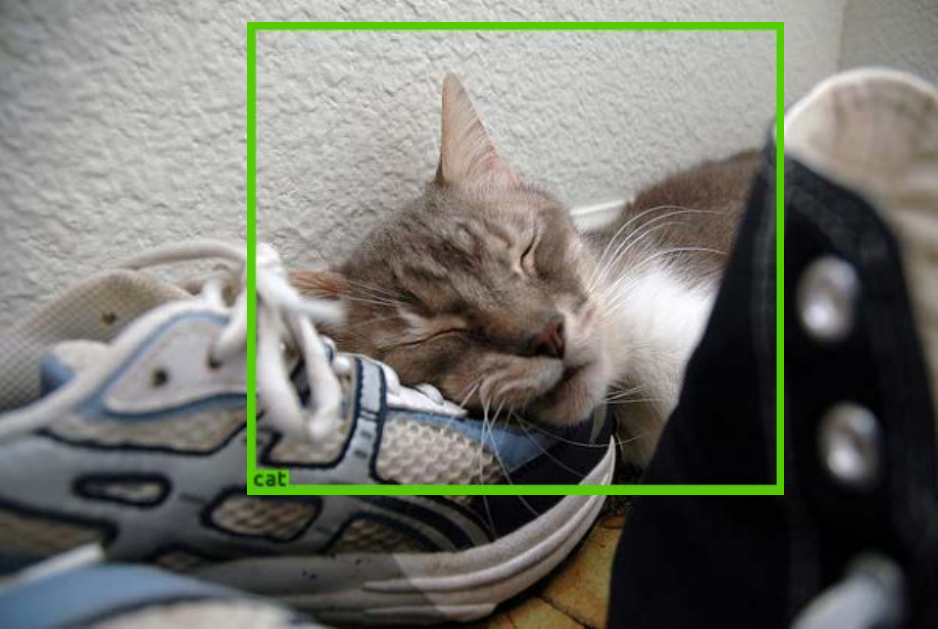}
&
\includegraphics[width=0.2\textwidth]{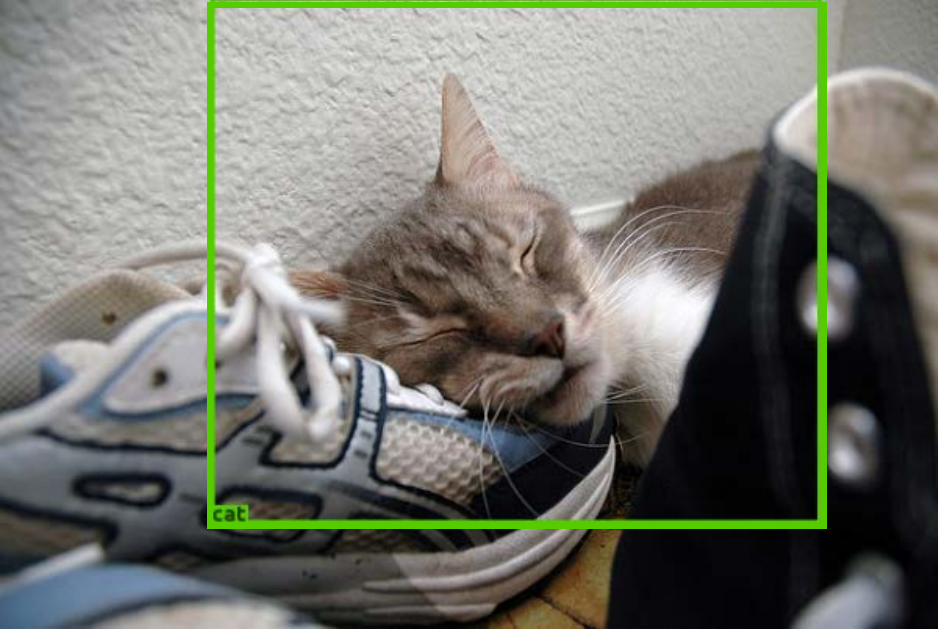}
\\
\includegraphics[width=0.2\textwidth]{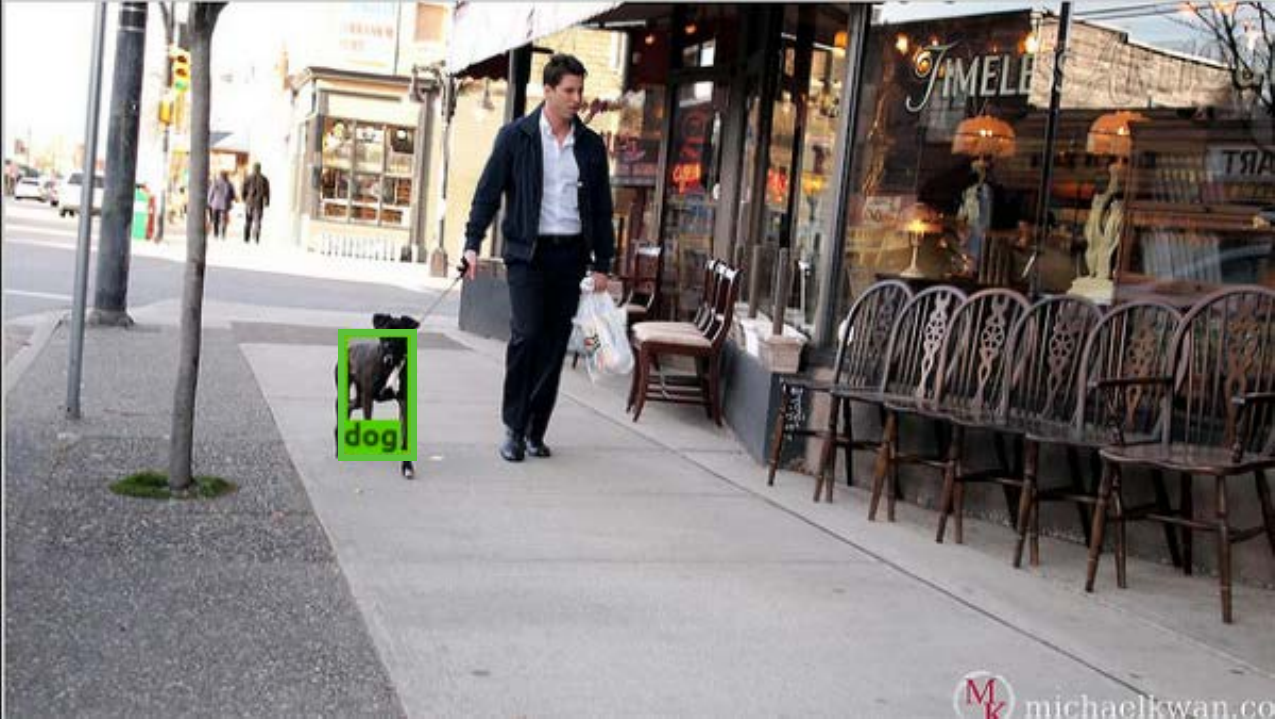} & \includegraphics[width=0.2\textwidth]{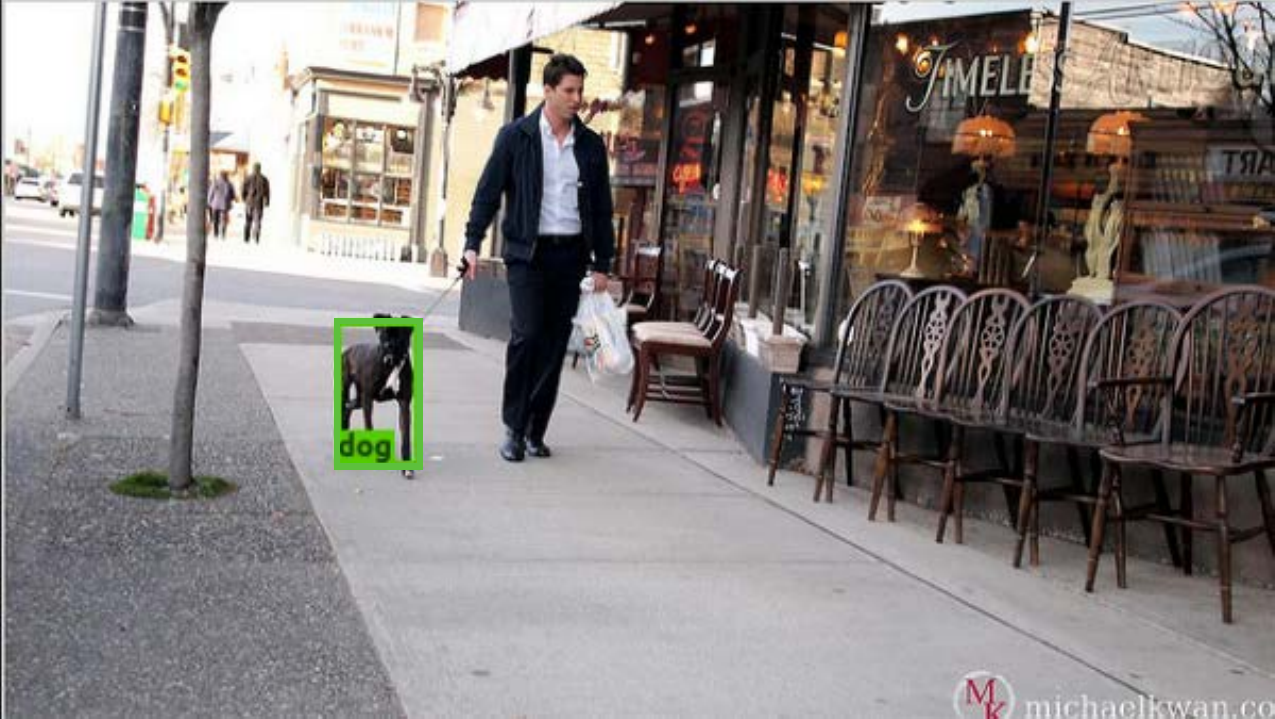} &
\includegraphics[width=0.2\textwidth]{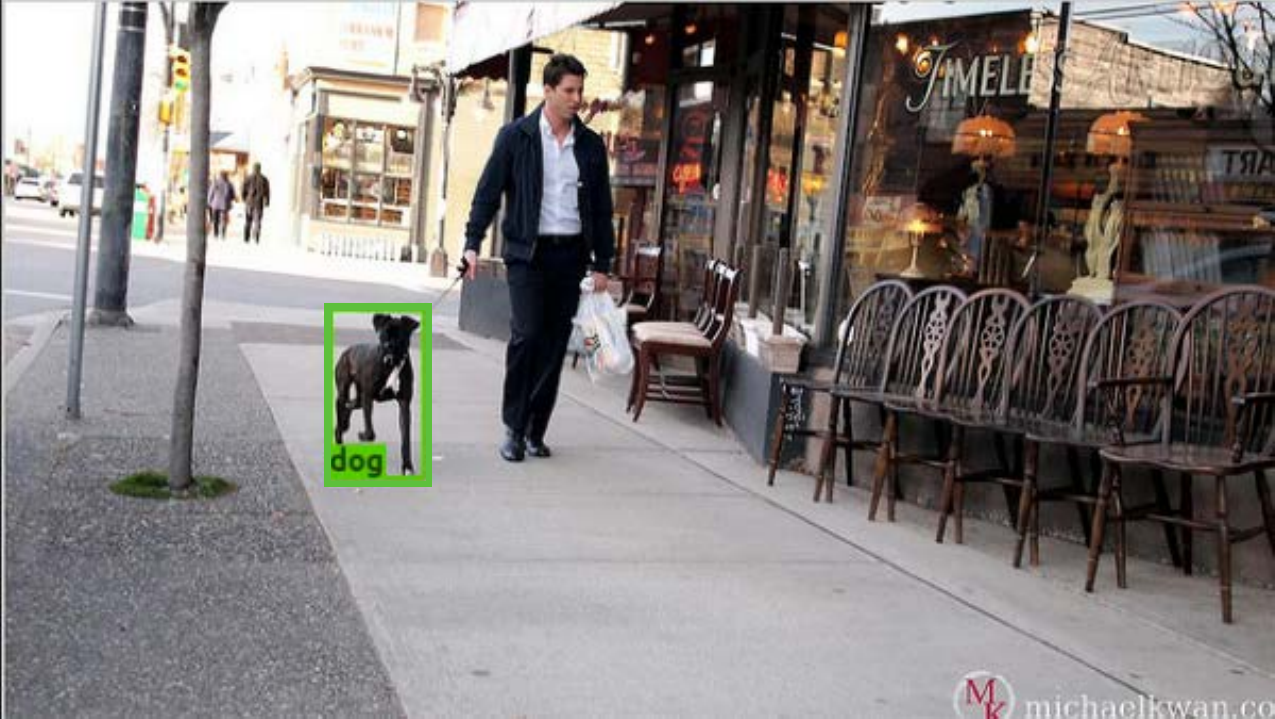} &
\includegraphics[width=0.2\textwidth]{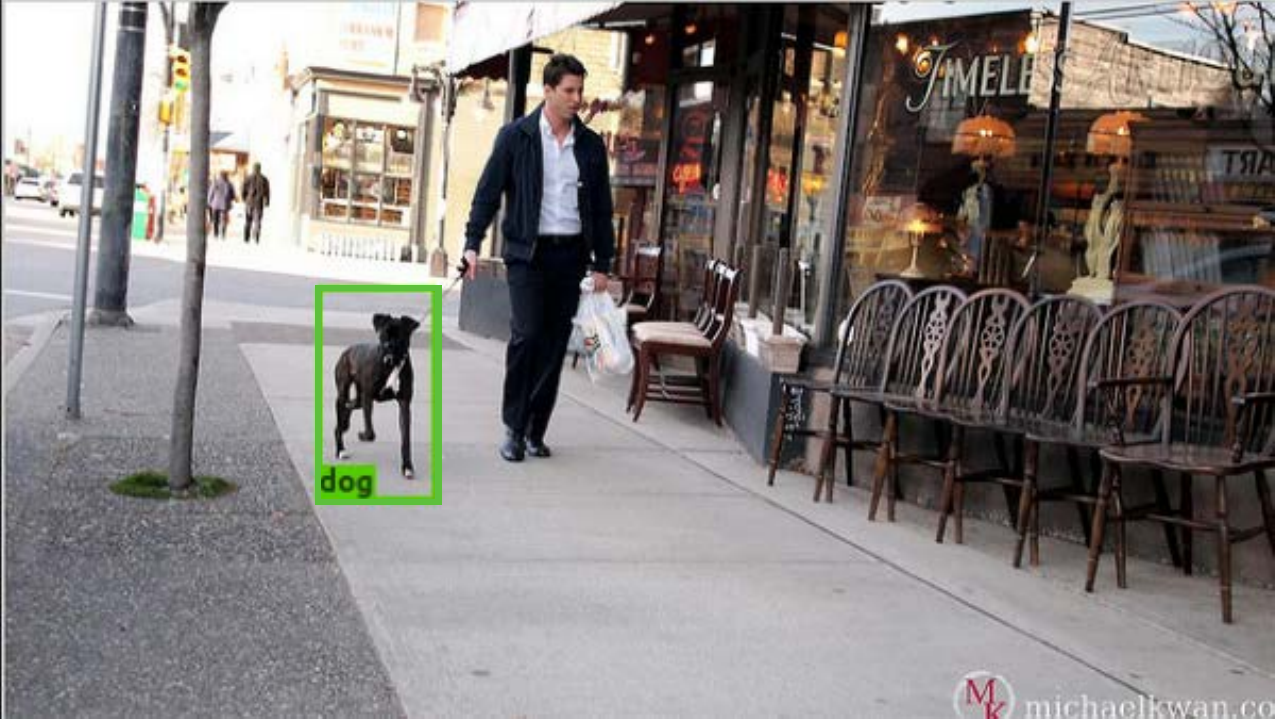} &
\includegraphics[width=0.2\textwidth]{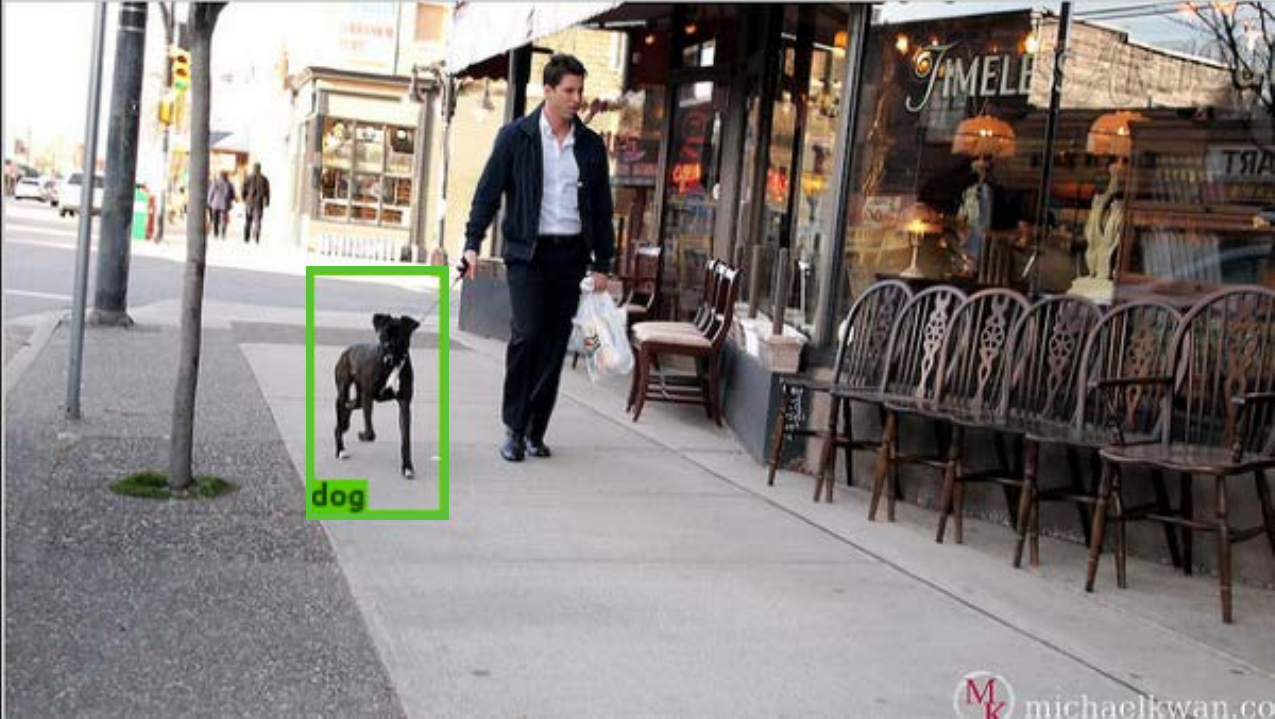} 
\\
\includegraphics[width=0.2\textwidth]{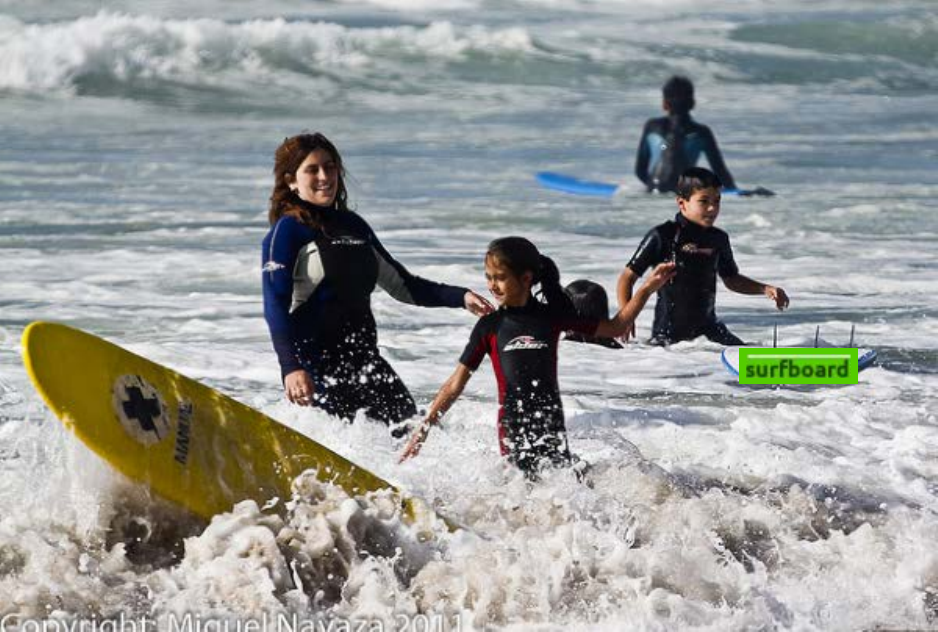} & \includegraphics[width=0.2\textwidth]{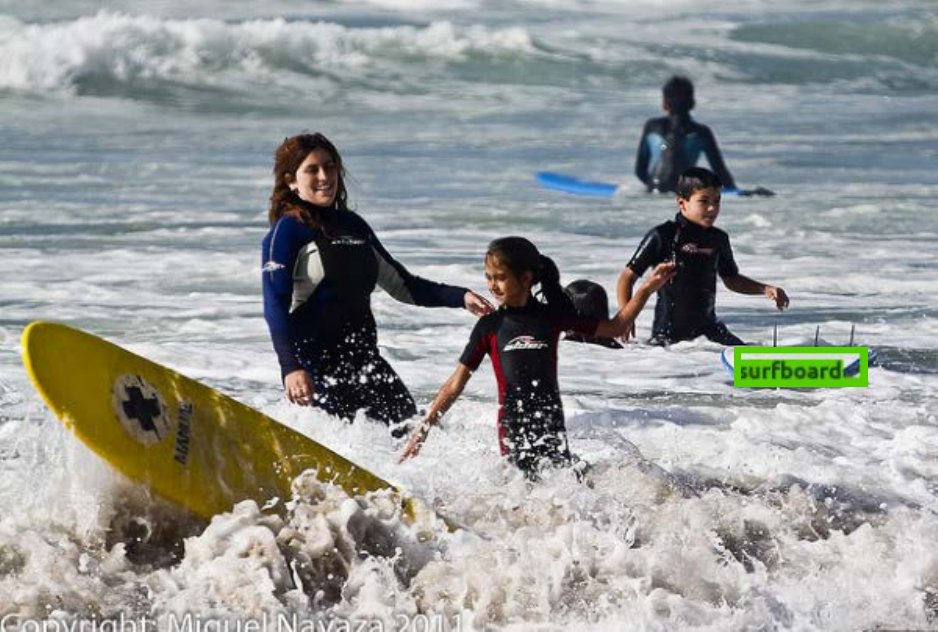} &
\includegraphics[width=0.2\textwidth]{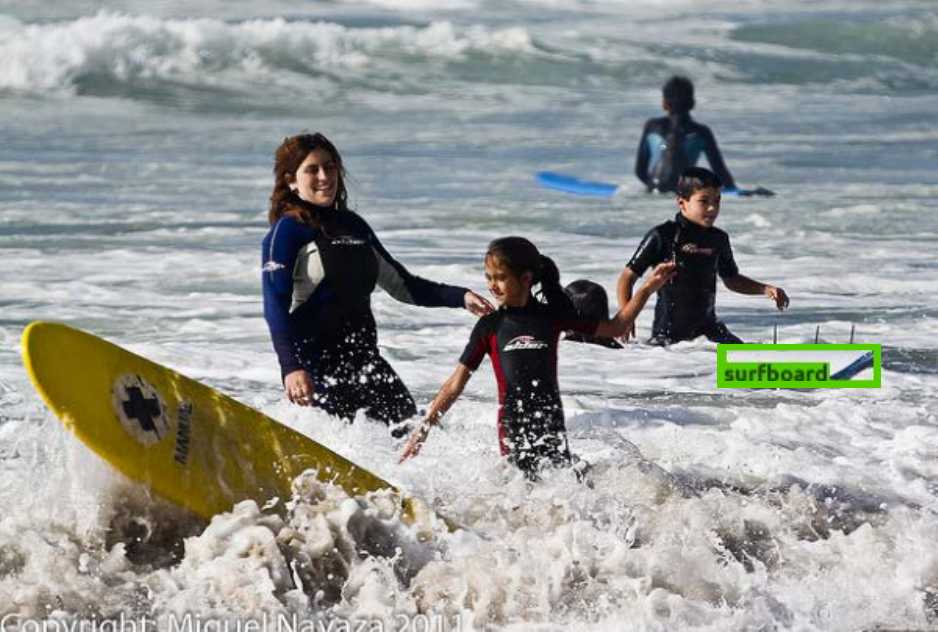} &
\includegraphics[width=0.2\textwidth]{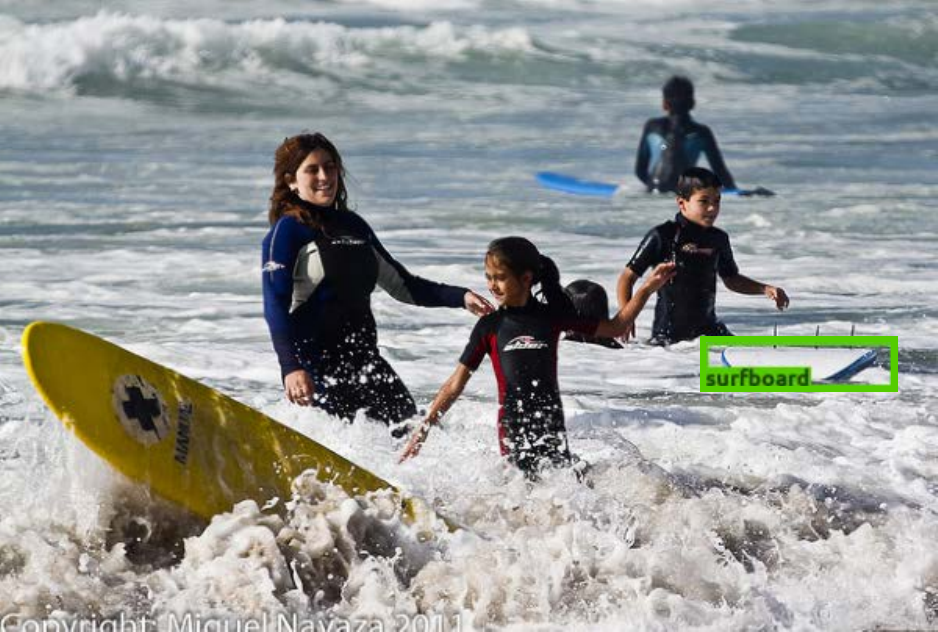} &
\includegraphics[width=0.2\textwidth]{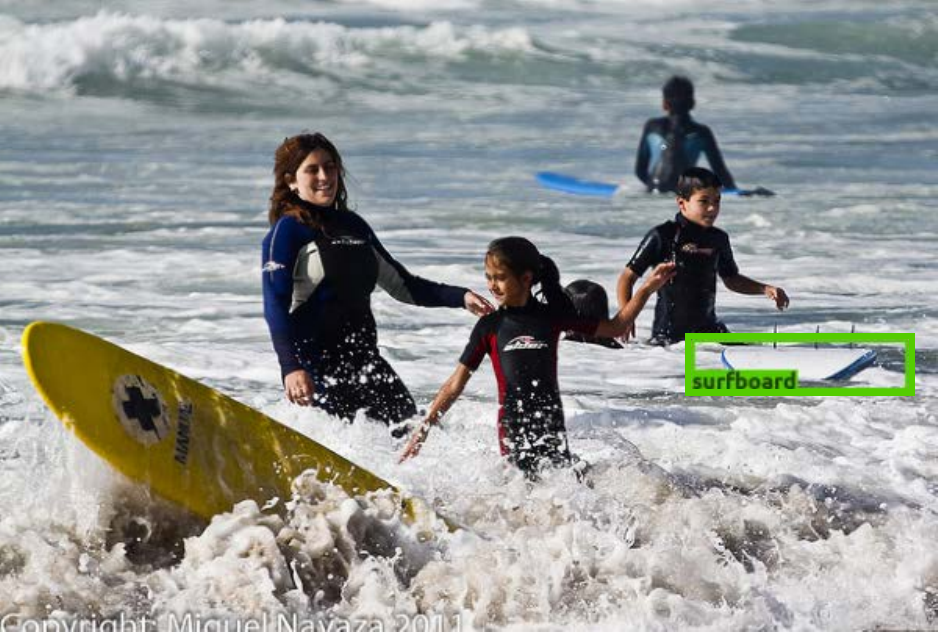} 
\\
Factor 0.5 & Factor 0.67 & Model prediction & Factor 1.5 & Factor 2.0 
\end{tabular}
\caption{
Scaling the model detections. Example of a bounding box predicted for a large object (first row), a medium object (second row) and a small object (third row) with Faster R-CNN (3rd column) and its scaled versions. In the left two images, the area of the bounding box is reduced by a scaling factor of, respectively, $0.5$ and $0.67$, whilst in the right two images the box area is increased by a factor of $1.5$ and $2$. 
}
\label{fig:scaling_example}
\end{figure*}

\begingroup
\setlength{\tabcolsep}{10pt}

\begin{table}[t]
\centering
\caption{AP (\emph{i.e.}, AP@$[0.5:0.95]$) and AP50 ($\%$) 
calculated for the predictions of three detectors 
on the MS COCO validation set and for the predicted boxes scaled with different scaling factors. 
Scaling the predicted boxes reduces the AP scores drastically.}
\label{tab:AP}
\fontsize{8}{10}\selectfont
\begin{tabular}{l|cc|cc|cc}
\toprule

Scaling factor & \multicolumn{2}{c}{Faster R-CNN} & \multicolumn{2}{c}{RetinaNet} & \multicolumn{2}{c}{Cascade R-CNN} \\
& AP & AP50 & AP & AP50 & AP & AP50 \\

\midrule
 0.50 & 0.0 & 0.1  & 0.1  & 0.4 & 0.0  & 0.1 \\
 0.67 & 5.1 & 37.1  & 5.4  & 38.2 & 5.6  & 40.7 \\
 1.00 & 36.7 & 54.1  & 37.4  & 56.7 &  39.6  & 53.7 \\
 1.50 & 5.5 & 38.4  & 6.2  & 41.3 & 5.7  & 41.3 \\
 2.00 & 0.0 & 0.2  & 0.3  & 1.3 & 0.0  & 0.2 \\
\bottomrule  
\end{tabular}
\end{table}

\endgroup

Given that, for the same IoU, humans prefer larger boxes and real object detectors do not tend to predict too large boxes, here we evaluate how humans judge re-scaled boxes. We do this through a user study, where we ask participants to evaluate five scaling factors, determined by scaling up or down the area of the predicted boxes with a factor of 1.5 and 2.0: $\{0.5, 0.67, 1.0, 1.5, 2.0 \}$. Large bounding boxes are cropped to not exceed the image boundaries. Examples 
of bounding box scaling for a large and a small object are 
shown in Figure~\ref{fig:scaling_example}. 
We refer to this study as \textit{Scaling Preference}. 
We ask the participants to choose the boxes they believe best identify a specific object in an image. The interface used in the user study 
allows the participants to 
select multiple options if they cannot determine a single best one.  
We use six random images selected from the MS COCO validation set 
per each combination between object size (\textit{small}, \textit{medium}, \textit{large}) 
and IoU range. We select 
five IoU ranges from $0.5\leq\text{IoU}<0.6$ to $0.9\leq\text{IoU}<1.0$ that correspond to true positive predictions, for a total of 90 images. 
We conduct this 
\textit{Scaling Preference} study on Faster R-CNN, RetinaNet, and Cascade Mask R-CNN.  

Scaling the detections of a well-performing object detector results in a slight change in appearance but a significant drop in AP. For a scaling of $1$, the baseline AP is $36.7\%$, yet a scaling of $1.5$ corresponds to a $\approx86\%$ decrease in AP. For a scaling factor of $2.0$, the AP is $\approx0\%$. 
Even with a more lenient IoU threshold, the AP50 decreases rapidly with both upscaling and downscaling.
As shown in Table \ref{tab:AP}, this behavior is consistent across the three object detectors.





\subsection{Results for the \textit{Scaling Preference} study}

\begin{figure}[]
\centering

\begin{tabular}{c@{}c}
\includegraphics[width=0.5\textwidth]{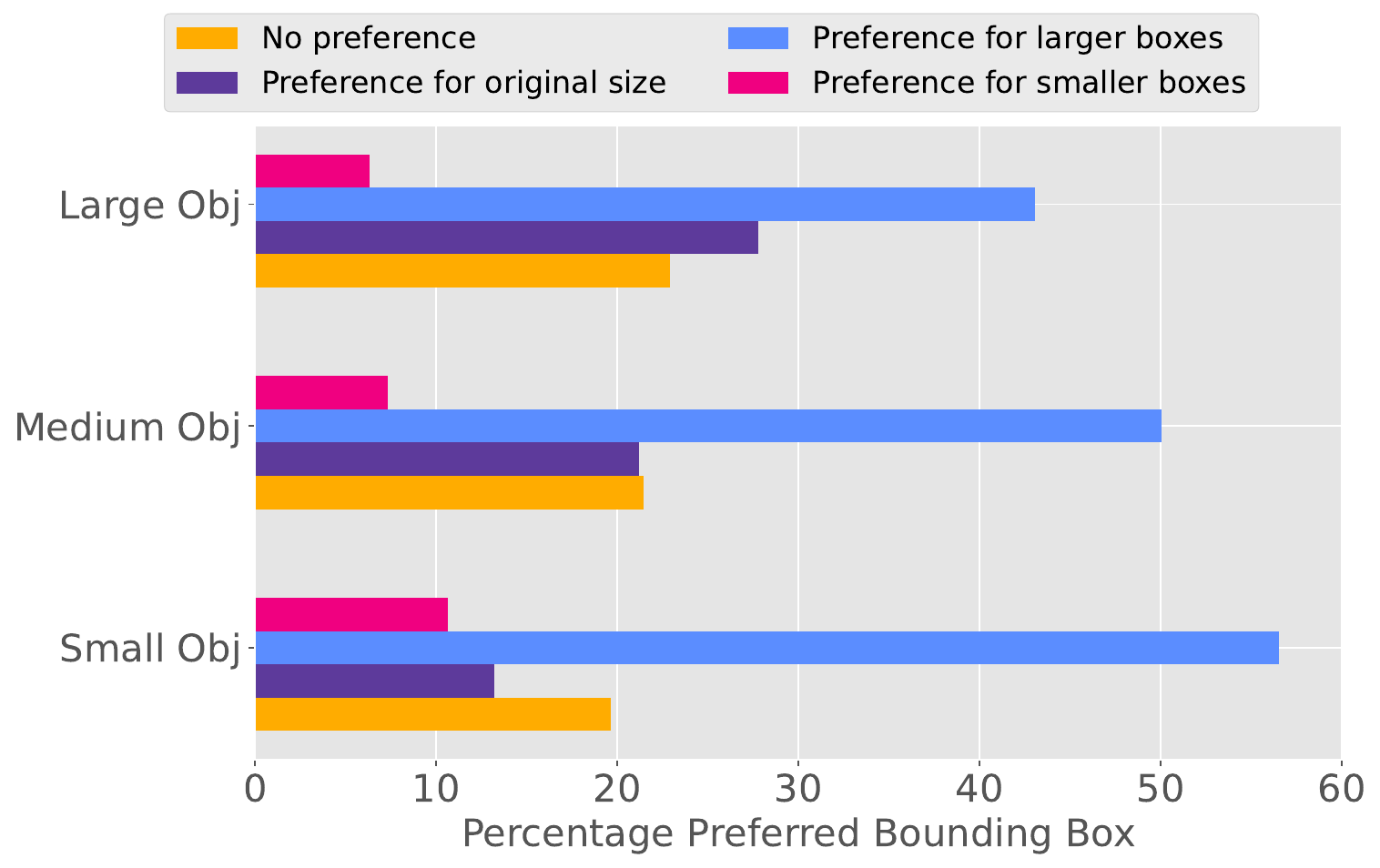} & 
\includegraphics[width=0.5\textwidth]{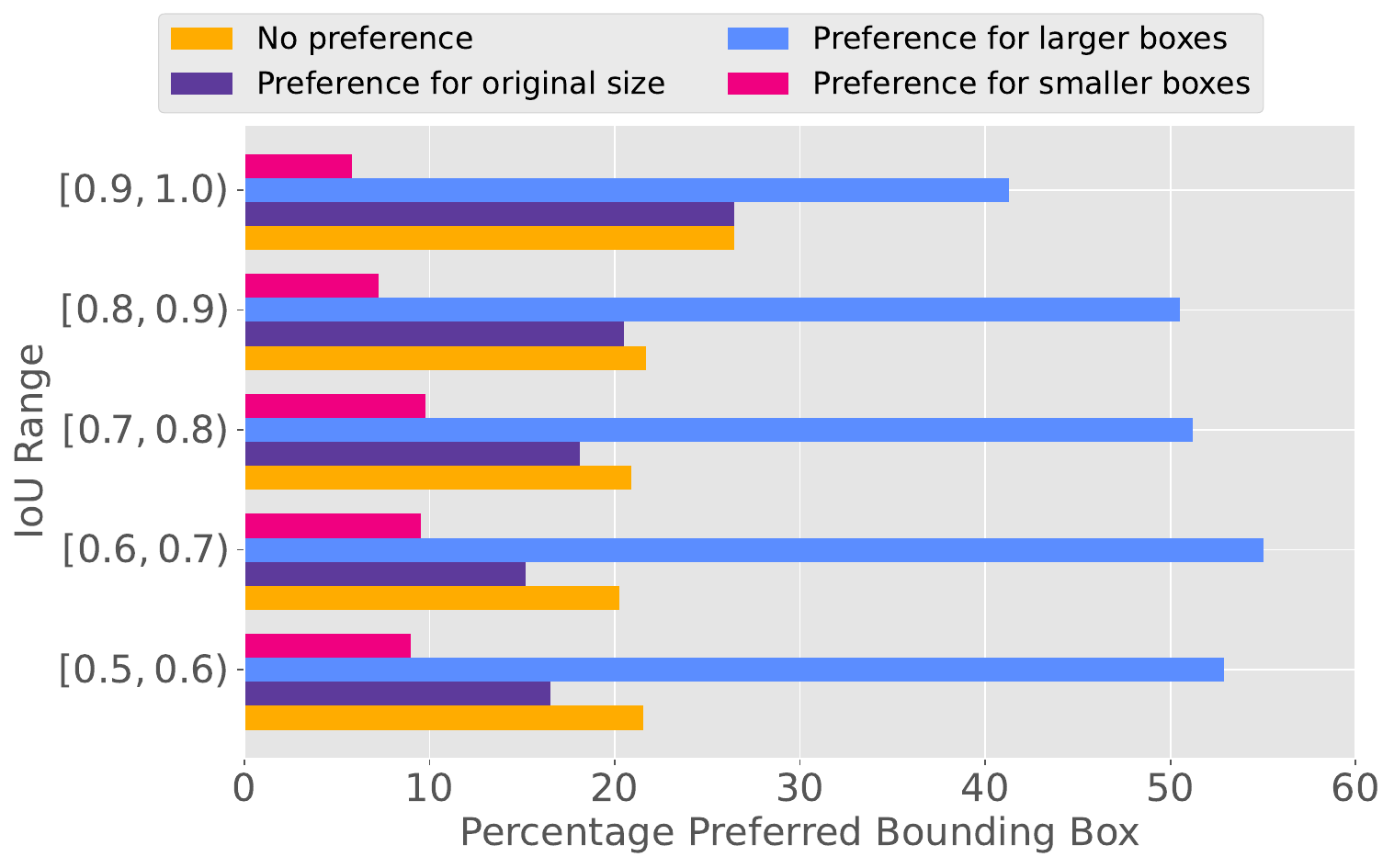} \\
\multicolumn{2}{c}{(a) Faster R-CNN}\\

\includegraphics[width=0.5\textwidth]{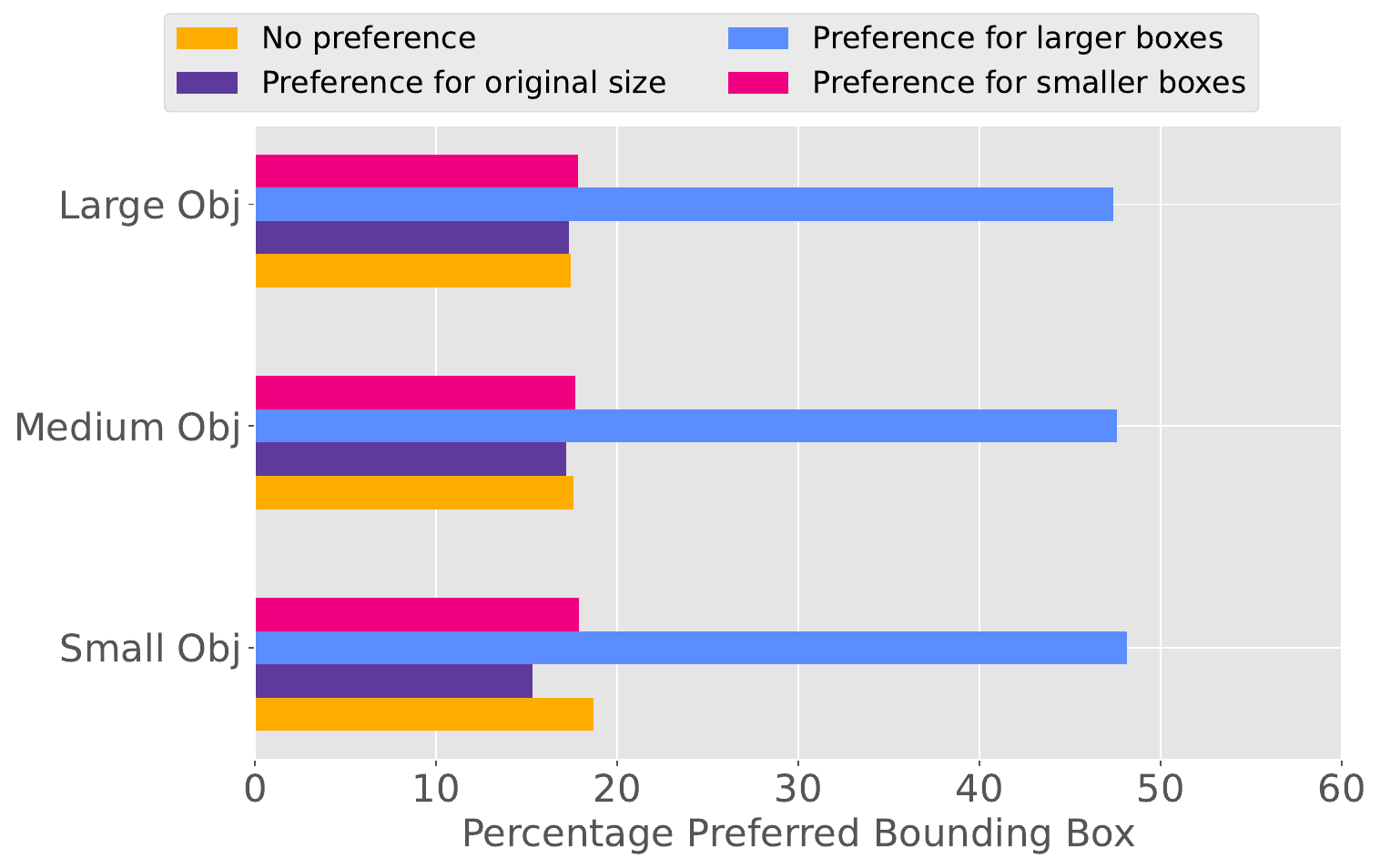} &
\includegraphics[width=0.5\textwidth]{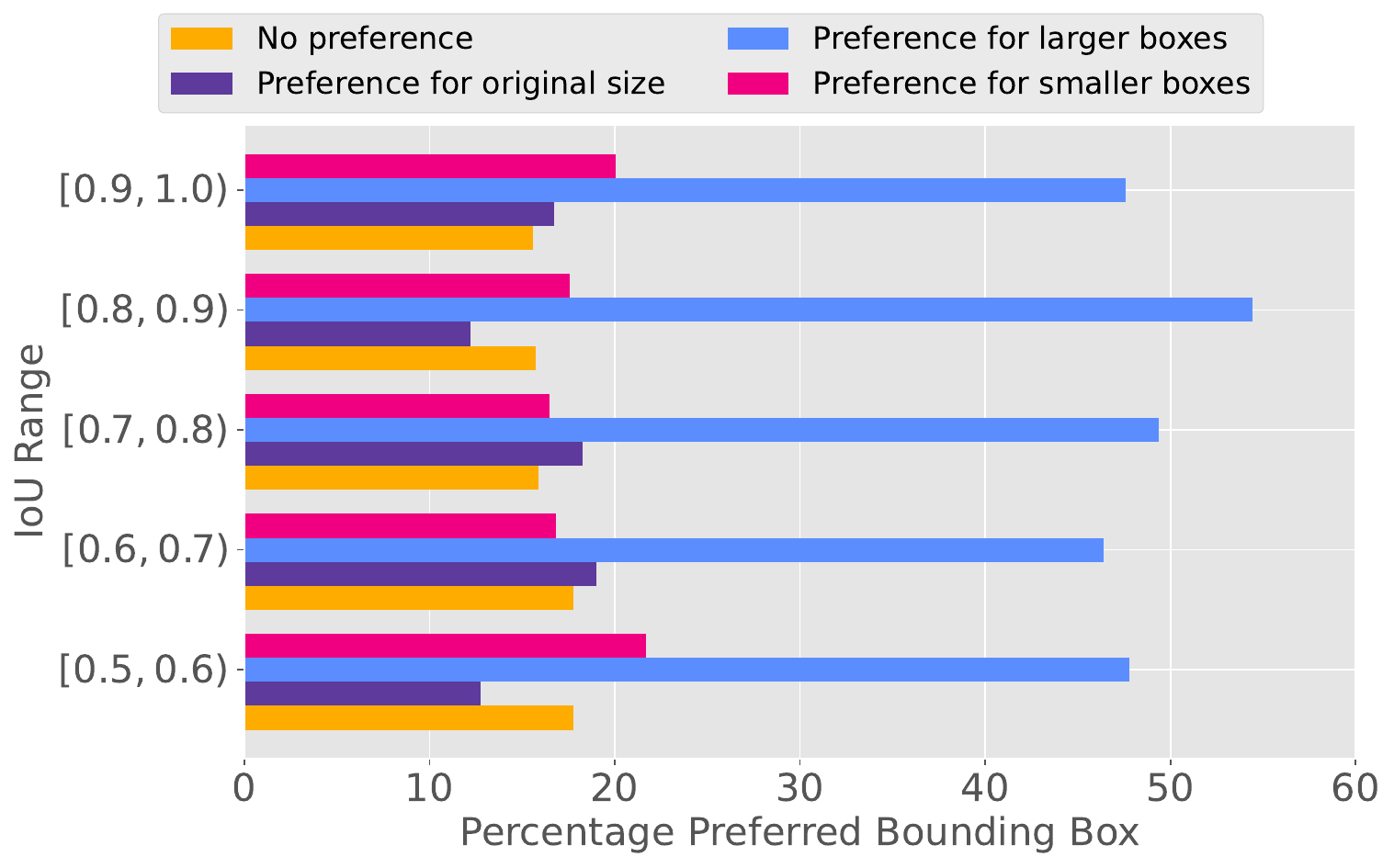} \\
\multicolumn{2}{c}{(b) RetinaNet}\\

\includegraphics[width=0.5\textwidth]{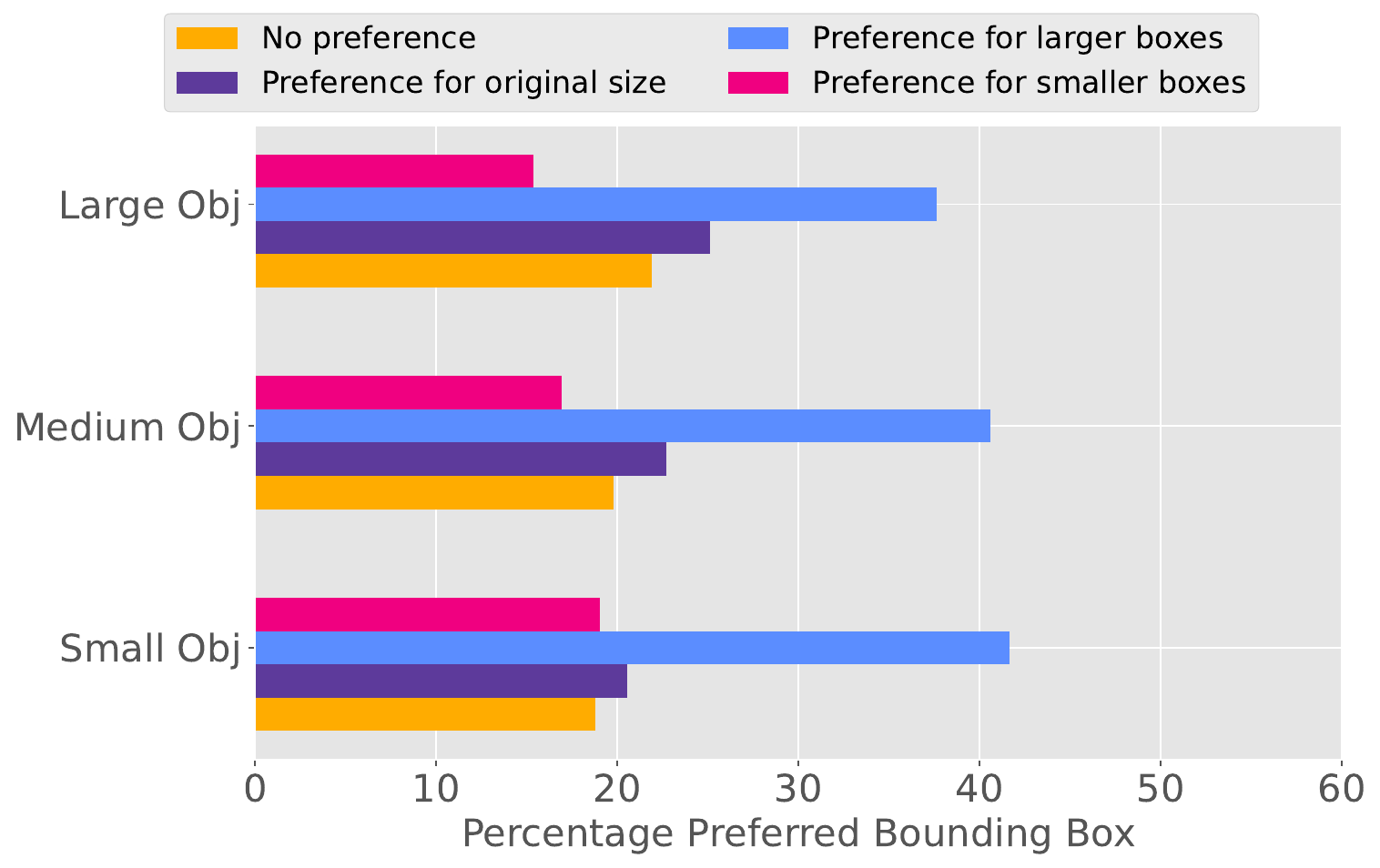} &
\includegraphics[width=0.5\textwidth]{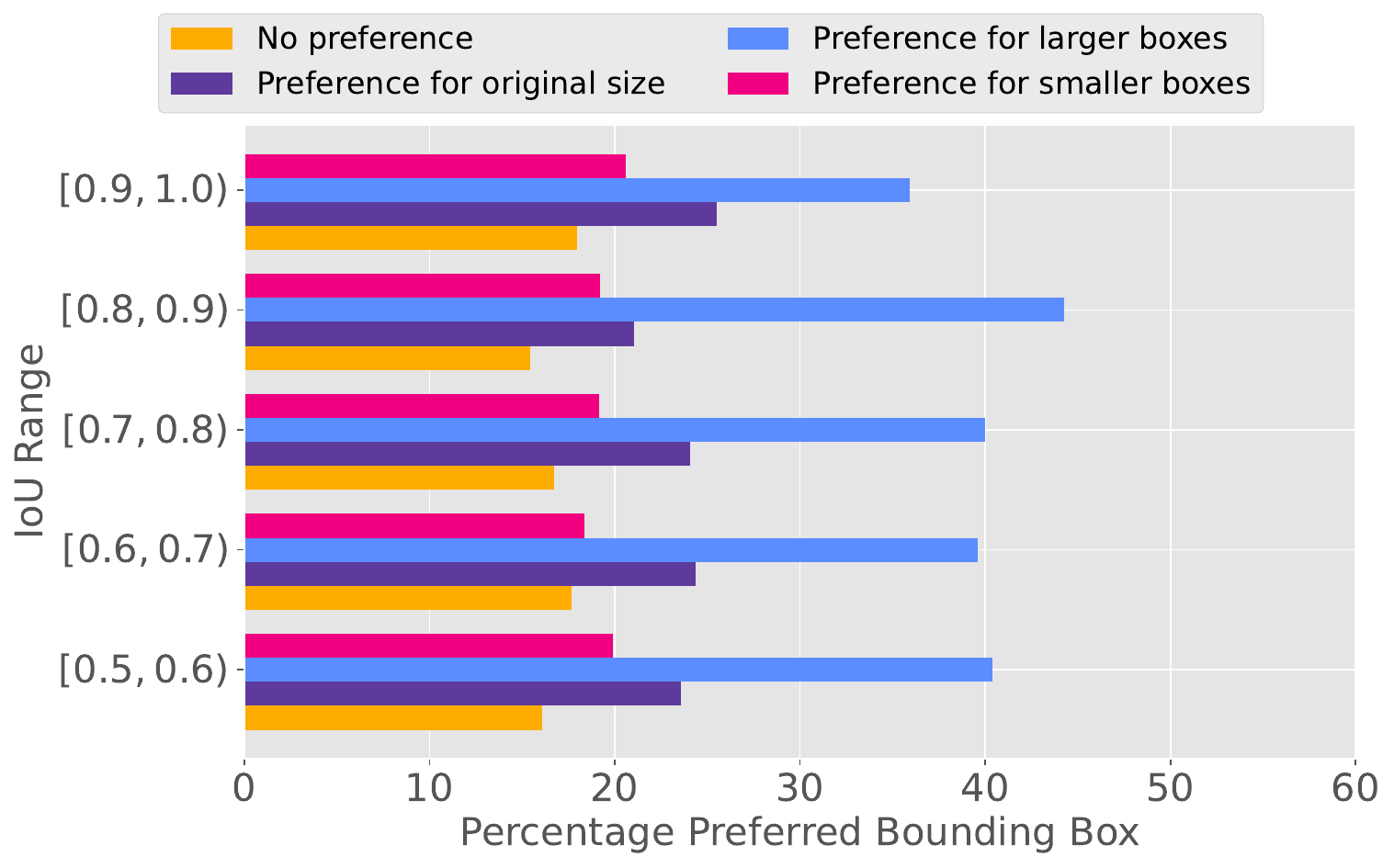}\\
\multicolumn{2}{c}{(c) Cascade Mask R-CNN}\\

\end{tabular}
\caption{
Results from the \emph{Scaling Preference} user study. The histograms show the percentage of preferred bounding box size 
per object category (S, M, L) and IoU range, from $0.5 \leq \text{IoU} < 0.6$ to $0.9\leq \text{IoU} < 1.0$, 
for three object detectors. 
The plots indicate that humans significantly prefer larger boxes.}
\label{fig:scaling_preference_study}
\end{figure}

Table \ref{tab:overview_scaling_preference} shows the number of participants and total number of judgments for the \textit{Scaling Preference} study. 
We use the Cochran’s  Q  test \cite{cochran1950comparison} to determine whether there are statistically significant differences in participants’ preferences regarding box sizes. In addition, we apply the posthoc Dunn tests with Bonferroni correction \cite{weisstein2004bonferroni} to find what are the scaling factors that result in significant differences in users' preferences. 
An overview of the results is provided in Figure \ref{fig:scaling_preference_study}. 
We group the scaling choices into (i) "Preference for smaller boxes", if a user selected the box scaled with factor $0.67$, the box scaled with factor $0.5$ or both; (ii) "Preference for larger boxes", if a user selected the box scaled with factor $1.5$, the box scaled with factor $2.0$ or both; (iii) "Preference for original size" if a user selected only the bounding box predicted by the model and (iv) "No preference" for all the remaining combinations of selections. Larger bounding boxes are consistently selected more often than small bounding boxes and than the original bounding box size for all three object detectors. This holds for 
different object sizes (Figure \ref{fig:scaling_preference_study}, left column), and IoU ranges (Figure \ref{fig:scaling_preference_study}, right column). 

Despite the preference for larger boxes, we cannot find a statistically significant difference between the preference for upscaling factor $1.5$ and upscaling factor $2.0$.
For Faster R-CNN, the preference for larger boxes is composed of $56.77\%$ of selections of both boxes scaled with factor $1.5$ and $2.0$; of $19.30\%$ of selections for scaling factor $1.5$ and of $23.93\%$ of selections for scaling factor $2.0$. 
Here, Dunn's test shows no statistically significant difference between the preference for scaling factor $1.5$ and scaling factor $2.0$. This means that larger boxes are preferred, but there is no single best upscaling factor. One exception holds for the bounding boxes for small objects predicted with Faster R-CNN: in this case, scaling factor $2.0$ is preferred over scaling factor $1.5$ (Dunn's $\alpha \approx 0$). This preference is an indicator that, for small objects, scaling the bounding box with a large scaling factor, like $2.0$, results in more satisfactory detections. A majority of votes for the largest box for small objects, albeit not statistically significant, is observed for the other object detectors.  
It is noticeable how larger bounding boxes are preferred to bounding boxes predicted with high IoU. This indicates that, for representative images of the diverse MS COCO dataset, humans are likely to prefer bounding boxes larger than the ground truth bounding boxes.  

\begingroup
\setlength{\tabcolsep}{10pt}

\begin{table}[]
\centering
\caption{Overview of participants and their judgments in the scaling preference study.}
\label{tab:overview_scaling_preference}
\fontsize{8}{10}\selectfont

    \begin{tabular}{lccc} \toprule
    \multicolumn{1}{c}{} &  Faster R-CNN & RetinaNet &  Cascade Mask R-CNN \\ \midrule
    Participants & 39 & 36 & 48 \\ 
    Judgments & 5400 & 5220 & 5632 \\
    \bottomrule
    \end{tabular}
\end{table}

\endgroup

\section{Asymmetric regression loss to encourage larger detections}
\label{sec:experiment2}
We find that humans consistently prefer larger object detections, while object detectors predict large and small boxes equally often. We propose an asymmetric bounding box regression loss that encourages larger detections. Our asymmetric loss is obtained by 
a simple modification of the smooth $L_1$ localization loss function used in standard object detectors. We use the asymmetry term $\alpha$ to increase the loss value when the predicted area is smaller than the ground truth area and decrease the loss value when is larger. 
The asymmetric loss is given by 
\begin{equation} 
{ \text{Asymmetric } L_{1,\text{smooth}}}= 
\begin{cases}
    \frac{1}{2\sqrt{\alpha}\beta}x^2, & \text{if } 0 \leq x < \beta\\
    \frac{\sqrt{\alpha}}{2\beta}x^2, & \text{if } -\beta < x < 0\\
    \frac{1}{\sqrt{\alpha}}x-\frac{\beta}{2\sqrt{\alpha}}, & \text{if } x \geq \beta\\
    -\sqrt{\alpha} x-\frac{\sqrt{\alpha}\beta}{2}, & \text{if } x \leq -\beta
\end{cases}
\end{equation}

\begin{figure}[]
\centering
  \includegraphics[width=0.8\columnwidth]{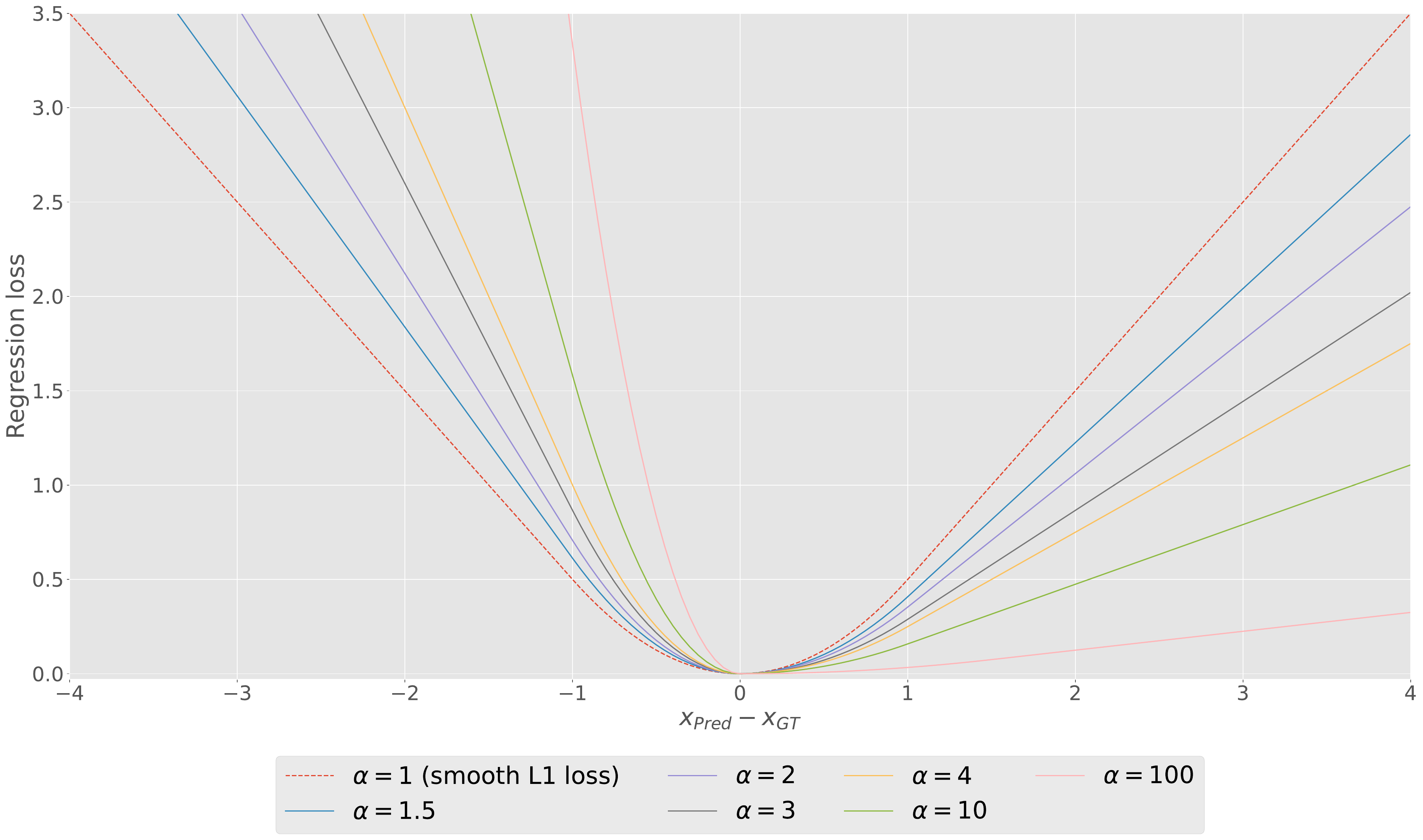}
  \caption{Asymmetric smooth $L_1$ loss with different $\alpha$. Larger bounding boxes are penalized less than smaller predicted boxes.}
  \label{fig:asymmetric_loss}
\end{figure}

The $\alpha$ represents the asymmetry term, $\beta$ determines the standard 
smoothing interval in which the $L_1$ loss becomes quadratic, and $x$ is the input to the loss function, which is the difference between the predicted height/width and the ground truth values, $x=x_{\text{pred}}-x_{\text{GT}}$. As shown in Figure~\ref{fig:asymmetric_loss}, the asymmetric loss is identical to the smooth $L_1$ loss when $\alpha=1$. We use the asymmetric loss function for the regression of the boxes' height and width. 

\begin{figure}[ht!]
\centering
  \includegraphics[width=0.8\columnwidth]{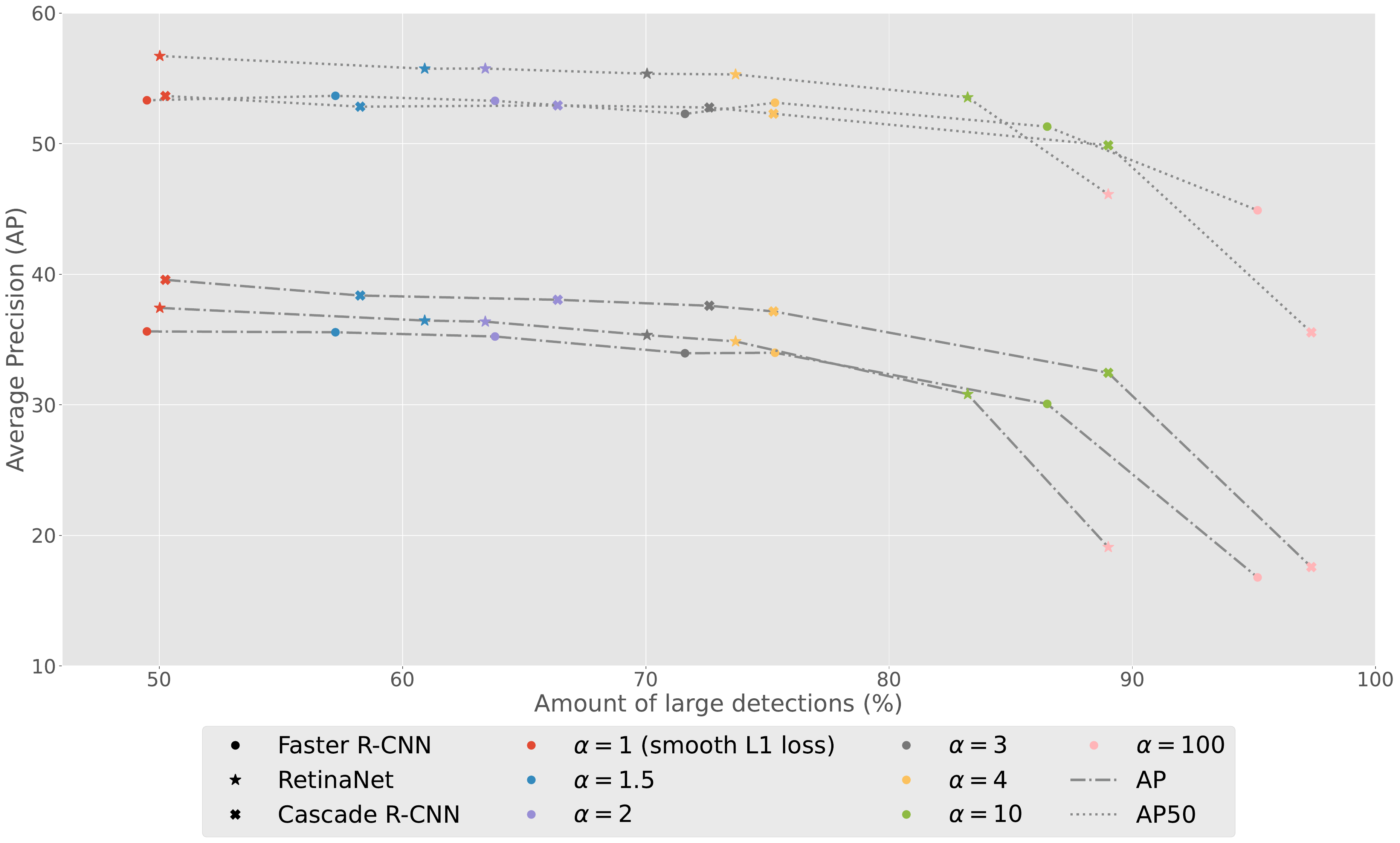} 
  \caption{Average Precision (AP) as a function of the amount of predicted boxes that are larger than the ground truth boxes. The percentage of large detections increases with the $\alpha$ parameter, while the AP decreases. 
  }
  
  \label{fig:AP_vs_large_boxes}
  \includegraphics[width=0.8\columnwidth]{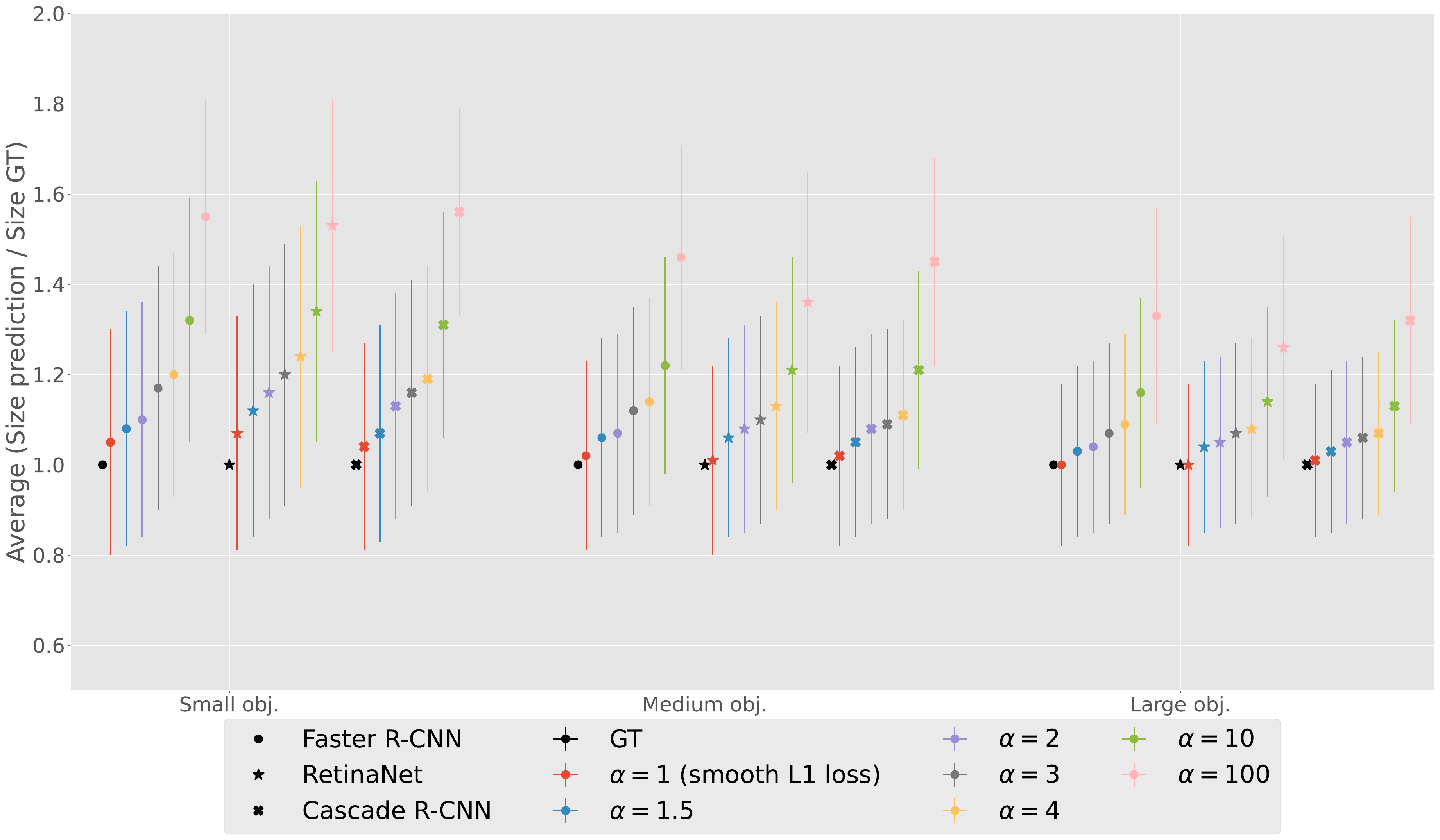}
  \caption{Bounding box size increases after fine-tuning object detectors with the asymmetric smooth $L_1$ loss with parameter $\alpha$.}
  \label{fig:box_size_increase}
\end{figure}

We fine-tune Faster R-CNN, RetinaNet, and Cascade R-CNN on MS COCO for 100k iterations with the asymmetric loss. As a result, the fine-tuned models are more likely to predict larger boxes over smaller boxes. Figure \ref{fig:AP_vs_large_boxes} shows the percentage of large detections for different $\alpha$ values. Similarly to the fixed scaling factors in the \textit{Scaling Preference} study in section~\ref{sec:experiment1}, we observe a decrease in the AP with the increase of large detections. With $\alpha=10$, we obtain 80\% to 90\% 
large predictions without compromising AP too much.

We measure the average size increase of the predicted bounding boxes compared to the ground truth. As shown in Figure~\ref{fig:box_size_increase}, increasing the $\alpha$ coefficient results in an increase of the average box size, for all three models and object sizes. The models fine-tuned with $\alpha=10$ return detections scaled compared to ground truth, 
on average by factors $1.21\pm0.24$ for Faster R-CNN, $1.21\pm0.25$ for RetinaNet, and $1.19\pm0.22$ for Cascade R-CNN, while fine-tuning with $\alpha=100$ results in average scaling of $1.41\pm0.26$ for Faster R-CNN, $1.34\pm0.28$ for RetinaNet, and $1.39\pm0.24$ for Cascade R-CNN. 
It is noticeable that the size of small objects' detections increases more than for medium and large objects. This is mostly due to small bounding boxes having more opportunity for expansion in the image, while large objects' boxes are already close to the image boundaries.

\subsection{Does the asymmetric $L_1$ loss lead to detections closer to the human preference?}
\label{sec:user_eval}
We conduct a final user study to investigate whether adopting the asymmetric $L_1$ loss results in detections closer to human preference.
In the evaluation study, we include the detections from the original pretrained Faster R-CNN ($\alpha=1$), the detections from Faster R-CNN fine-tuned with $\alpha$ 10 and 100, and the detections scaled by a fixed factor 1.5, which was one of the preferred options in the \textit{Scaling Preference} study~\ref{sec:experiment1}. 
These values for $\alpha$ are chosen to have detection sizes that notably differ from the Faster R-CNN baseline (Figure~\ref{fig:box_size_increase}).
We ask users to compare the four different detections for the same object and choose the one that, in their opinion, best identifies the object. We include 45 detections, equally sampled from the three object categories (\textit{small}, \textit{medium}, \textit{large}). We conduct the study on Amazon Mechanical Turk \cite{mturk} and collect 660 judgments. 

The results are summarized in Table~\ref{tab:user_study_results}. The Cochran's Q test reveals statistically significant differences between the proportions of preferred object detections. The detections obtained by fine-tuning with asymmetric loss, $\alpha=10$, are always the most preferred. This preference is statistically significant when considering all object categories and small objects (Dunn's $\alpha \leq 0.001$). In the other cases, fine-tuning with $\alpha=10$ is significantly more preferred than scaling with a fixed factor (Dunn's $\alpha \approx 0$), 
thus confirming the advantage of the asymmetric loss over fixed scaling. 

The preference for the asymmetric loss over fixed scaling is likely due to the fixed scaling factor upscaling all boxes equally, irrespective of the object size. Conversely, using the asymmetric loss results in boxes upscaled more for small objects than for medium and large objects, as illustrated in Figure~\ref{fig:box_size_increase}. This might lead to higher human preference, since large objects are already easily identifiable with a tighter box. In fact, as shown in Table~\ref{tab:user_study_results}, the fixed scaling 1.5 is almost never chosen for large objects. 
In addition, we observe that the most preferred option --- asymmetric loss with $\alpha=10$ --- leads to detections that are, on average, larger than the ground truth by a factor between 1.1 and 1.5. We hypothesize that the optimal scaling factor might lie within this range.
Another potential reason why scaling by 1.5  is less preferred is that the fixed scaling strategy retains the aspect ratio of the original predicted box. This aspect ratio may not be optimal when upscaling the boxes. In contrast, the asymmetric loss function imposes fewer constraints on the aspect ratio.  

 %
%
%
Overall, fine-tuning the models with our asymmetric $L_1$ loss results in detections closer to human preference. We suggest adopting this loss when object detections are meant to be presented to humans.

\begingroup
\setlength{\tabcolsep}{8pt}

\begin{table}[]
\centering
\caption{Users' preferred object detections (\%), computed with Faster R-CNN fine-tuned with the asymmetric loss function or up-scaled with factor 1.5. Fine-tuning with $\alpha=10$ is always the most preferred option.}
\label{tab:user_study_results}
\fontsize{8}{10}\selectfont
\begin{tabular}{llcccc}
\toprule
\multirow{2}{*}{Object cat.} & \multirow{2}{*}{\# judgments} & \multicolumn{4}{c}{Chosen object detection (\%)} \\ \cmidrule{3-6}
& & $\alpha=1$ & $\alpha=10$ & $\alpha=100$ & Scal. fact. 1.5 \\ 
\midrule

\textit{All} & 660 & 27.4 & 37.1 & 21.2 & 14.2          \\
\textit{Small} & 229 & 17.0 & 31.9 & 26.6 & 24.5        \\
\textit{Medium} & 215 & 27.9 & 39.5 & 16.7 & 15.8       \\
\textit{Large} & 216 & 38.0 & 40.3 & 19.9 & 1.9          \\
\bottomrule  
\end{tabular}%
\end{table}

\endgroup


\subsection{Qualitative analysis of the preferred boxes}
\label{sec:qualitative_analysis}

\begin{figure*}[]
	\centering
	\frame{\includegraphics[width=1\textwidth]{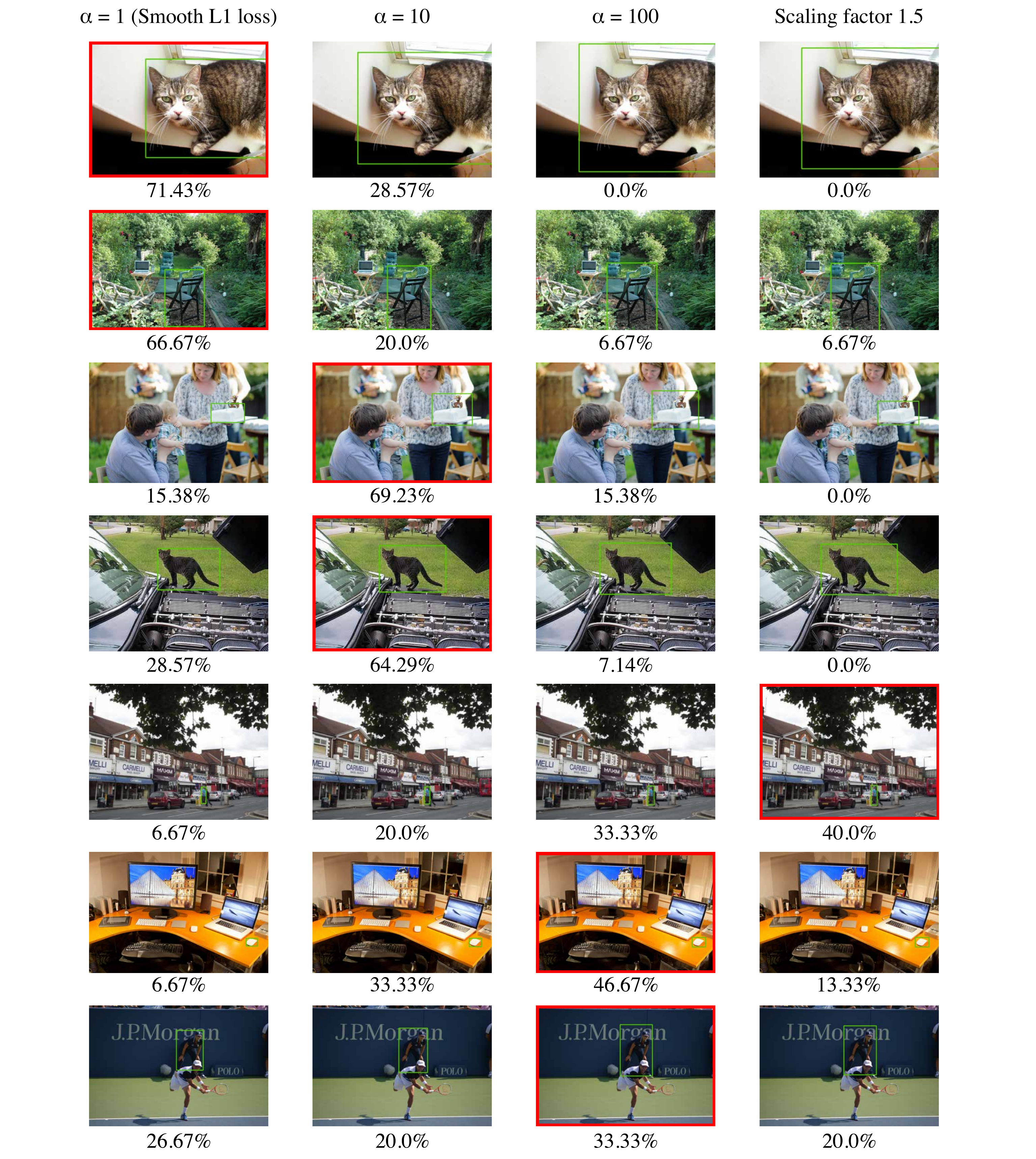}}
	\caption{Example of human preferences obtained from the user evaluation of the asymmetric loss. The columns show the percentage of users who prefer the bounding boxes obtained by the original pretrained Faster R-CNN ($\alpha=1$), after fine-tuned with $\alpha$ 10 and 100, or scaling by a fixed factor 1.5. Generally, humans prefer tight boxes for large objects (first row) and when the object of interest overlaps with other objects (second row).  
    Slightly larger boxes, obtained with asymmetric loss, $\alpha = 10$ or $100$, are preferred when small parts of the object protrude outside too tight bounding boxes (e.g., the candle on the birthday cake, third row), or partly covered by the box line itself (fourth row).
    Large boxes ($\alpha=1$ or scaling factor 1.5) are chosen for very small objects (fourth and fifth row).
    Finally, we found no preference when all bounding boxes are too visually similar (last row). 
    }
	\label{fig:qualitative_eval}
\end{figure*}

We manually analyze the results obtained from the user evaluation of the asymmetric loss and illustrate some representative examples in Figure \ref{fig:qualitative_eval}. We notice that the tight bounding boxes predicted by the Faster R-CNN baseline, namely, trained with $\alpha=1$, are generally preferred for large objects, e.g., the \textit{cat} in the first row. Preference for $\alpha=1$ also occurs when there are multiple objects behind or in the proximity of the object of interest. In the image on row 2 of Figure \ref{fig:qualitative_eval}, 
larger bounding boxes partly include the \textit{chair} behind the one of interest. In this case, tight boxes delineate better the subject of focus.

Slightly larger boxes, obtained by fine-tuning with our asymmetric loss, $\alpha=10$, are preferred when small parts of the object are not contained in the tight bounding box predicted by Faster R-CNN (e.g., 
the \textit{candle} on the birthday cake in Figure \ref{fig:qualitative_eval}, row 3), or partly covered by the box contour itself, like the 
ears and tail of the \textit{cat} in row 4.
We hypothesize that predicted tight bounding boxes 
leave out possible object protrusions, despite resulting in high AP. 
Presumably, humans prefer large boxes because they can include the whole object. Additionally, in the presence of a uniform background (e.g., the green grass behind the cat in Figure \ref{fig:qualitative_eval}, row 4), humans are generally less concerned if the bounding box is slightly larger. Similarly, the asymmetric loss makes it more likely to include all the small protruding parts of the objects in the predicted boxes.

We observe that the preference for larger boxes, obtained by scaling with factor 1.5 or with the asymmetric loss $\alpha=100$ occurs when the objects of interest are very small, e.g., the \textit{mouse} and the \textit{person} walking on the street (row 5 and 6, Figure \ref{fig:qualitative_eval}). 
Finally, in a few cases, the original Faster R-CNN detector, the detectors fine-tuned with the asymmetric loss or manually scaled result in very similar boxes, indistinguishable by a human eye. In this situation, we observe no clear human preference, as for the \textit{person} in the last row in Figure~\ref{fig:qualitative_eval}.

The qualitative analysis suggests that there exists a relationship between the object characteristics, especially size (already observed in \cite{strafforello2022humans}) and shape, and the preferred bounding box size. We leave the investigation of the factors that determine the user preference for future work.


\section{Conclusion}
\label{sec:discussion}
%


Prior work \cite{strafforello2022humans} shows that humans prefer larger boxes in a fully controlled setup. In this paper, we confirm this result in practice, with real detectors. We evaluate the bounding boxes predicted by three popular object detectors. 
We find that the object detectors predict large and small bounding boxes equally often, therefore are not aligned with the human preference found in \cite{strafforello2022humans}.
In addition, humans consistently prefer larger bounding boxes over the predicted boxes, even with AP approximately zero. 
Therefore, we recommend being careful with AP scores when object detectors are intended for human use: a high AP does not automatically correspond to high human preference.   

It is noticeable how the preference occurs even for bounding boxes predicted with high IoUs: this suggests that humans are likely to prefer larger bounding boxes compared to tight ground truth bounding boxes. 

We propose an asymmetric loss function that encourages detectors to predict large boxes more often than small boxes, without having to re-annotate the training images. Our user evaluation shows that fine-tuning with the asymmetric loss results in object detections more aligned with human preference. 
%
After qualitatively analyzing the results collected from our study, we hypothesize that the human preference is affected by the object characteristics, such as shape and size. For example, generally tight boxes are preferred for large objects, while larger boxes are preferred for small objects. Further investigation into these observations may be considered in the future. 

\section*{Acknowledgements}
This work is part of the research program Efficient Deep Learning (EDL), which is (partly) financed by the Dutch Research Council (NWO). 

%
%
\bibliographystyle{splncs04}
\bibliography{main}
\end{document}